%% file: paper.tex
\documentclass[]{fairmeta}
\usepackage{times}
\usepackage{latexsym}

\usepackage[T1]{fontenc}
\usepackage[utf8]{inputenc}

\usepackage{microtype}

\usepackage{inconsolata}

\usepackage{graphicx}
\usepackage{array}
\usepackage{tabularx}
\usepackage{amsmath}
\usepackage{amssymb}
\usepackage{hyperref}
\usepackage{enumitem}
\usepackage{geometry}
\usepackage{natbib}
\usepackage{tcolorbox}
\usepackage{booktabs}
\usepackage{multirow}
\usepackage[table]{xcolor}
\definecolor{lightgreen}{RGB}{230,255,230}
\usepackage{makecell}
\usepackage{threeparttable}
\usepackage{adjustbox}
\usepackage{bm}
\usepackage{pifont} % For checkmarks and crosses if needed, or use math symbols
\newcommand{\cmark}{\ding{51}}%
\newcommand{\xmark}{\ding{55}}%

\newcommand{\squishlist}{
   \begin{list}{$\bullet$}{%
        \setlength{\itemsep}{0pt}%
        \setlength{\parsep}{0pt}%
        \setlength{\topsep}{0pt}%
        \setlength{\partopsep}{0pt}%
        \setlength{\listparindent}{-2pt}%
        \setlength{\itemindent}{-5pt}%
        \setlength{\leftmargin}{1.2em}%
        \setlength{\labelwidth}{0em}%
        \setlength{\labelsep}{0.5em}%
    }
}
\newcommand{\squishend}{
    \end{list}  }

\newcommand{\sys}{MobileLLM-Flash}

\title{MobileLLM-Flash: Latency-Guided On-Device LLM Design for Industry Scale Deployment}

\author{
Hanxian Huang$^{*}$, Igor Fedorov$^{*}$, Andrey Gromov, Bernard Beckerman, Naveen Suda, David Eriksson, Maximilian Balandat, Rylan Conway, Patrick Huber, Chinnadhurai Sankar, Ayushi Dalmia, Zechun Liu, Lemeng Wu, Tarek Elgamal, Adithya Sagar, Vikas Chandra, Raghuraman Krishnamoorthi
}

\contribution[*]{Co-first authors}
\affiliation[]{Meta AI}
\abstract{Real-time AI experiences call for on-device large language models (OD-LLMs) optimized for efficient deployment on resource-constrained hardware. The most useful OD-LLMs produce near-real-time responses and exhibit broad hardware compatibility, maximizing user reach. We present a methodology for designing such models using hardware-in-the-loop architecture search under mobile latency constraints. This system is amenable to industry-scale deployment: it generates models deployable without custom kernels and compatible with standard mobile runtimes like Executorch. Our methodology avoids specialized attention mechanisms and instead uses attention skipping for long-context acceleration. 

Our approach jointly optimizes model architecture (layers, dimensions) and attention pattern. To efficiently evaluate candidates, we treat each as a pruned version of a pretrained backbone with inherited weights, thereby achieving high accuracy with minimal continued pretraining. We leverage the low cost of latency evaluation in a staged process: learning an accurate latency model first, then searching for the Pareto-frontier across latency and quality.

This yields \sys{}, a family of foundation models (350M, 650M, 1.4B) for efficient on-device use with strong capabilities, supporting up to 8k context length. \sys{} delivers up to \bm{$1.8\times$} and \bm{$1.6\times$} faster prefill and decode on mobile CPUs with comparable or superior quality. 
Our analysis of Pareto-frontier design choices offers actionable principles for OD-LLM design.}

\date{\today}
\correspondence{hanxianhuang@meta.com, ifedorov@meta.com}

\begin{document}

\maketitle
    \vspace{-1em}
\begin{figure}[!h]
    \centering
    \includegraphics[width=0.6\linewidth]{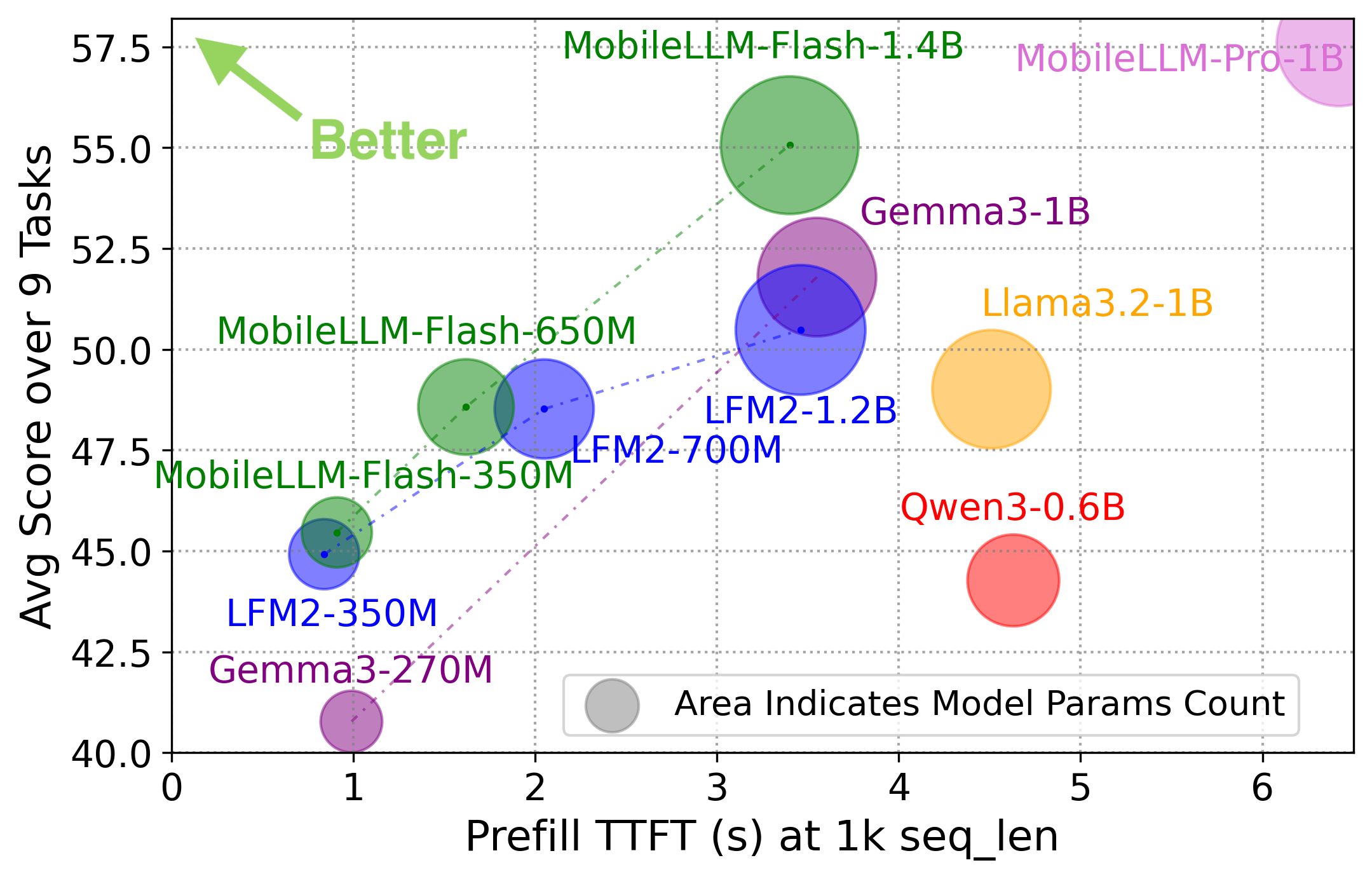}
    \vspace{-1em}
    \caption{Comparison between \sys{} and state-of-the-art OD-LLMs. \sys{} achieves up to \bm{$1.8\times / 1.6\times$} faster prefill/decode on mobile CPUs with superior accuracy than LFM2~\cite{LFM2}. Evaluation details are in Sec.~\ref{sec:experimental-results}.}
    \label{fig:fig1_comparison}
    \vspace{-1em}
\end{figure}

\input{sections/1-intro}
\input{sections/2-related}

\input{sections/3-architecture-search}
\input{sections/4-results}

\section{Conclusion}
We introduce \sys{}, a model family optimized for mobile latency through a novel 2-stage hardware-in-the-loop architecture search. By prioritizing shallow-and-wide structures and interleaved skip-attention patterns, we achieve state-of-the-art quality with significant speedups over strong baselines. Our methodology is compatible with Executorch and establishes a practical, scalable framework for delivering efficient, real-time AI on edge devices without specialized kernels.

\section*{Limitations}
Due to the high computational cost of training candidate architectures, our Bayesian Optimization search focused exclusively on architectural parameters (depth, width, attention patterns). We did not perform a co-optimization of training hyperparameters (e.g., learning rate schedules, optimizer settings) via Ax. It is possible that specific architectural candidates could achieve higher quality with bespoke hyperparameter tuning, which was outside the scope of this study.

To ensure immediate industry-scale deployability and compatibility with standard runtimes like ExecuTorch, in this paper we did not explore novel sub-quadratic attention mechanisms (such as SSMs or linear attention variants mentioned in Sec.~\ref{sec:related}) that currently lack mature runtime support. Extending the search space to include these emerging architectures remains a direction for future work as their software support matures.

\section*{Ethical Considerations}
Our work contributes to "Green AI" by focusing on efficiency. By optimizing for lower latency and smaller model sizes, MobileLLM-Flash reduces the computational energy required for inference. Furthermore, our pruning-based search method is data-efficient, requiring significantly fewer (only 35\%) training tokens (and thus less GPU energy) to discover optimal architectures compared to training candidates from scratch.

\clearpage
\newpage
\bibliographystyle{assets/plainnat}
\bibliography{paper}

\clearpage
\newpage
\beginappendix

\section{Efficiency Proxies} 
\label{sec:appendix-efficiency-proxy}

\begin{figure*}[!h]
    \centering
    \includegraphics[width=1\linewidth]{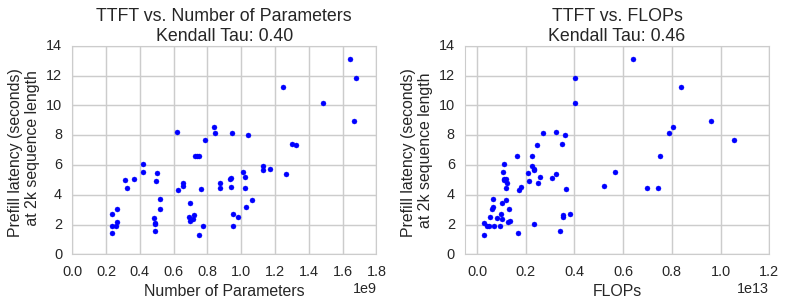}
    \caption{Kendall Tau correlation between TTFT at 2k sequence length vs. model parameter count and FLOPs.}
    \label{fig:kendall_tau1}
\end{figure*}

\begin{figure*}[!h]
    \centering
    \includegraphics[width=1\linewidth]{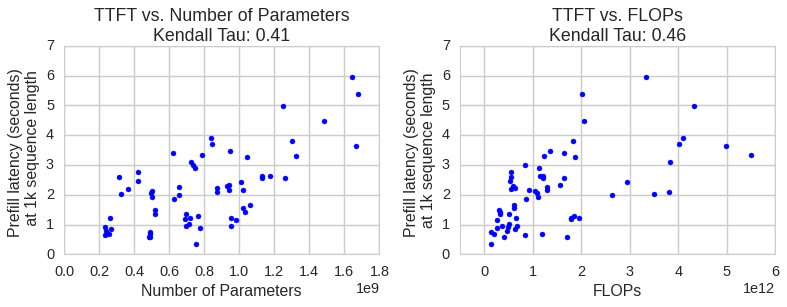}
    \caption{Kendall Tau correlation between TTFT at 1k sequence length vs. model parameter count and FLOPs.}
    \label{fig:kendall_tau2}
\end{figure*}

\begin{figure*}[!h]
    \centering
    \includegraphics[width=1\linewidth]{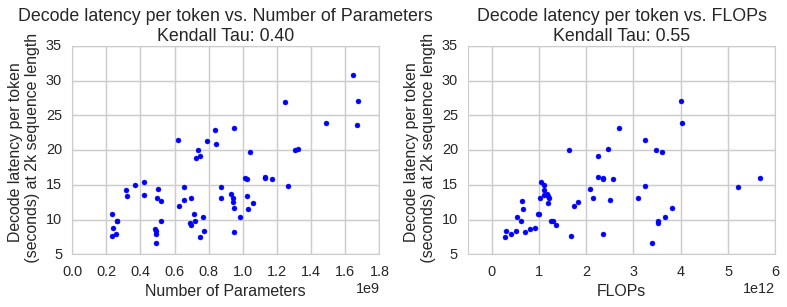}
    \caption{Kendall Tau correlation between decode latency at 2k sequence length vs. model parameter count and FLOPs.}
    \label{fig:kendall_tau3}
\end{figure*}

\begin{figure*}[!h]
    \centering
    \includegraphics[width=1\linewidth]{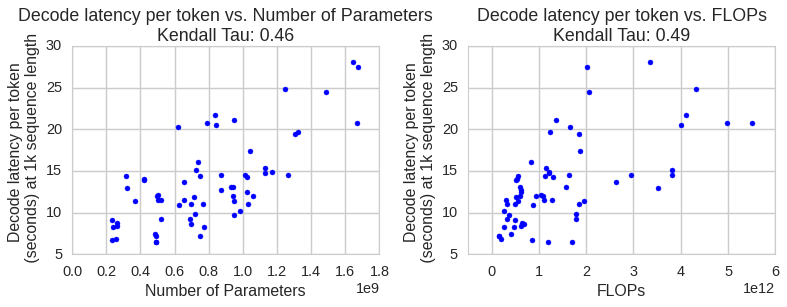}
    \caption{Kendall Tau correlation between decode latency at 1k sequence length vs. model parameter count and FLOPs.}
    \label{fig:kendall_tau4}
\end{figure*}
We present visualization of correlation coefficients between real measured prefill / decode speed vs. parameter count / FLOPs in Fig.~\ref{fig:kendall_tau1}, Fig.~\ref{fig:kendall_tau2},Fig.~\ref{fig:kendall_tau3}, and Fig.~\ref{fig:kendall_tau4} for various context lengths.

\section{Explanation for the inefficiency of SWA}\label{inefficient-SWA} 
A high level explanation for the inefficiency of SWA in our settings is as follows. We find that a prefill chunk size of 1024 produces the lowest time-to-first-token (TTFT), making the best use of parallelism across tokens in the context window. At the same time, Executorch constrains the SWA to be greater than or equal to the prefill chunk size, such that the sliding window is lower-bounded by $1024$. Therefore, when prefilling $2$k tokens, only the second chunk benefits from the SWA. At the same time, the ring-buffer implementation of SWA in Executorch requires computation of the entire attention matrix, as opposed to just the lower triangular portion as done during standard attention computation. As a result, in the $2$k prefill, $1024$ chunk size setting, we find that the benefits of SWA are outweighed by the drawback of the slower attention calculation and the model overall achieves worse TTFT with SWA.

\end{document}

%% file: sections/1-intro.tex
\section{Introduction}
% \vspace{-0.5em}
% \begin{figure}[!ht]
%     \centering
%     \includegraphics[width=\linewidth]{figures/ACL_comparison1.png}
%     % \vspace{-1.5em}
%     \caption{Comparison between \sys{} and state-of-the-art OD-LLMs. \sys{} achieves up to \bm{$1.8\times / 1.6\times$} faster prefill / decode on mobile CPUs with superior accuracy than LFM2~\cite{LFM2}. Evaluation / benchmarking details are in Sec.~\ref{sec:experimental-results}.}
%     \label{fig:fig1_comparison}
%     % \vspace{-1.5em}
% \end{figure}

Deployment of efficient on-device large language models (OD-LLMs) on resource-constrained devices, e.g., mobile phones and smart glasses, is critical for enabling real-time AI experiences. The two distinguishing features of real-world on-device AI assistants serving large scale industry-grade traffic are: 1) They must operate \textbf{near-real-time}, with a specific emphasis on time-to-first-token (TTFT, i.e., time from receiving the request to outputting the first generated token) since the decode rate can often be hidden by streaming the model output back to the user; 2) They must be close to \textbf{general-purpose} from a runtime perspective, avoiding complex building blocks that require specialized kernels. These requirements ensure models operate across diverse software and hardware stacks (Android, iPhone, wearables). 

The real-time requirement places an upper bound on TTFT, and therefore, on the number of prefill tokens. %Prefill latency is the time from receiving the input prompt to generating the first output token, dominated by processing all input tokens in parallel~\cite{fast-odllm}. 
While a $4$s TTFT can still yield a reasonable user experience, a $10$s TTFT does not~\cite{nielsen1994usability, kim2026seconds}. As such, models should be optimized for the most practical use cases. 
%the regime in which they are most likely to be useful. 
We find that $\sim2$k tokens is a practical sweet spot where OD-LLMs can both perform useful tasks and achieve TTFT $\leq 4$s (Tab.~\ref{tab:model_performance}). 

Many existing efficient LLM designs often optimize proxy metrics, such as parameter count or floating-point operations per second (FLOPs) rather than measured on-device latency, or focus on server-side optimizations~\cite{fu2025nemotronflash, gu2025jetnemotron, yang2025zebrallama}. 
However, these methods do not adequately capture the practical constraints and performance bottlenecks encountered in real-world on-device deployment, leaving a gap for methodologies grounded in empirical latency and hardware-in-the-loop evaluation. Furthermore, some designs relying on specialized kernels for efficient attention~\cite{gu2025jetnemotron,yang2025gated,mamba2} face barriers to large-scale adoption due to limited portability.

\noindent\textbf{Our main contributions are:}
\squishlist
% \noindent\textbf{1)} 
\item To our knowledge, the first joint architecture and attention pattern NAS for OD-LLMs, directly optimizing mobile CPU latency via a novel two-stage Bayesian optimization flow.
\item A pruning-based search that inherits pretrained weights, using only $35\%$ of the training tokens required by from-scratch training, making our hardware-in-the-loop LLM NAS practical.% at LLM scale. 
% We present a hardware-in-the-loop, pruning-based architecture search that directly optimizes mobile prefill latency. This yields deployment-ready models, compatible with Executorch.
% \noindent\textbf{2)} 
\item \sys{} ($350$M, $650$M, $1.4$B) (Fig.~\ref{fig:fig1_comparison}), a family of deployment-ready OD-LLMs with up to $1.8\times$ prefill and $1.6\times$ decode speedup on mobile CPUs, runnable out of the box without specialized kernels. 
%featuring a compact hybrid backbone with skip attention and grouped query attention. This yields significant speedups in prefill and decode on mobile CPUs and runs out of the box without specialized kernels.
% \noindent\textbf{3)} 
\item We present an analysis of the latency-accuracy Pareto frontier, deriving actionable design principles to guide future OD-LLM development. Our search reveals that interleaved skip-attention consistently outperforms sliding-window attention on mobile CPUs. 
\squishend

%% file: sections/2-related.tex
\vspace{-0.5em}
\section{Related Work}\label{sec:related}
% \vspace{-0.5em}
\textbf{OD-LLM design and architecture search.} 
%Prior works like MobileLLM~\cite{liu2024mobilellm}, MobileLLM-Pro~\cite{mobilellmpro}, and TinyLlama~\cite{zhang2024tinyllama} have demonstrated success in designing parameter-efficient LLMs within a given parameter budget. However, parameter efficiency alone often does not translate into proportional latency reductions on real devices. For example, previous LLMs~\cite{liu2024mobilellm, mobilellmpro} often adopt deep-and-thin model structures, which may result in suboptimal latency-accuracy trade-offs. 
While prior works~\cite{liu2024mobilellm,mobilellmpro} achieve parameter efficiency, their deep-and-thin structures often fail to improve on-device latency. 
Recent studies~\cite{acun2025composers, gu2025jetnemotron, fu2025nemotronflash, yang2025zebrallama, cowsik2025towards} use neural architecture search (NAS) to discover OD-LLMs, but focus on either parameter efficiency or server-side GPU optimization. 
LFM2~\cite{LFM2} introduces an edge-optimized model family based on a new block gated short convolution attention, but the underlying \texttt{conv1d} operator can perform poorly at small batch sizes or sequence lengths~\cite{heinecke2017understanding}. 
%At a larger scale, Puzzle~\cite{bercovich2025puzzle} applies distillation-based NAS to optimize 50--70B LLMs for server GPU throughput, but does not search attention patterns or target mobile CPU latency. 
%Latency-aware NAS has also been explored in other domains: NanoSD~\cite{sanyal2026nanosdedgeefficientfoundation} searches U-Net block types for on-device diffusion-based image restoration, but does not address LLM-specific dimensions such as attention patterns or the high cost of LLM candidate evaluation. 
%In contrast, our work jointly searches architecture shape and attention pattern for on-device scale LLMs under mobile latency constraints. 
Latency-aware NAS has been explored for server-scale 50--70B LLMs (Puzzle~\cite{bercovich2025puzzle}) and diffusion models (NanoSD~\cite{sanyal2026nanosdedgeefficientfoundation}), but neither searches attention patterns nor targets mobile CPU latency for on-device scale LLMs, which is the focus of our work.

\noindent\textbf{Efficient attention alternatives and hybrid models.} %Efficient attention modules with sub-quadratic complexity, such as Mamba2~\cite{mamba2}, DeltaNet~\cite{yang2024parallelizing}, Gated DeltaNet~\cite{yang2025gated} and JetBlock~\cite{gu2025jetnemotron}, have been proposed to reduce computation and memory costs. Despite their potential, these blocks are built for server-side GPUs, and do not currently have deployment support in deployment stacks such as Executorch, making them incompatible with any existing on-device runtime out of the box.  
Sub-quadratic attention modules like Mamba2~\cite{mamba2}, Gated DeltaNet~\cite{yang2025gated} and JetBlock~\cite{gu2025jetnemotron}, built for server-side GPUs, reduce compute and memory costs but lack on-device runtime support (e.g., Executorch).
%Hybrid models combining linear and quadratic attentions for improved recall and reasoning capabilities have been further explored. They often stack Mamba and Llama layers~\cite{lenz2025jamba,glorioso2024zambacompact7bssm,ren2025samba} or mixing RNNs/convolutions with attention~\cite{pilault2023blockstate,de2024griffinmixinggatedlinear,yang2025gated}. These approaches typically rely on manual operator selection, involving tedious trial-and-error. 
Hybrid models~\cite{lenz2025jamba,glorioso2024zambacompact7bssm,ren2025samba, pilault2023blockstate,de2024griffinmixinggatedlinear,yang2025gated} combines linear and quadratic attentions to improve recall and reasoning, but they typically rely on tedious manual design. 
Our work automates the selection of efficient attention patterns in hybrid models, utilizing runtime-supported mechanisms such as skip attention and sliding-window attention (SWA), enabling more scalable development and optimal design choices. 
Automatic search with sub-quadratic modules is performed in \citet{LFM2}; however, it does not adopt such hybrids, possibly due to the absence of highly optimized kernels they automatically lose by latency. %mysteriously rejected all such hybrids. Possibly because in 
A comparison of our paper with the most relevant related works is shown in Tab.~\ref{tab:related-work-comparison}.

\begin{table}[t!]
\centering
\caption{Comparison of methodologies.}
\label{tab:related-work-comparison}
\resizebox{\columnwidth}{!}{%
\begin{tabular}{lccc}
\toprule
\textbf{Feature} & \textbf{MobileLLM-Flash} & \makecell{\textbf{LFM2} \\ \cite{LFM2}} & \makecell{\textbf{Jet-Nemotron}~\cite{gu2025jetnemotron},\\ \textbf{Nemotron-Flash}~\cite{fu2025nemotronflash}} \\
\midrule Directly optimizes mobile CPU latency & \cmark &\cmark & \xmark \\
\midrule
Unified architecture (shape + attention) search 
    & \cmark & \xmark & \xmark\\
\midrule
Efficient pruning-based search 
    & \cmark & \xmark& \xmark \\
\midrule
Compatible with any hardware / runtime 
    & \cmark & \cmark & \xmark \\
\bottomrule
\end{tabular}%
}
\end{table}

\noindent\textbf{Bayesian optimization (BO)} is a sample-efficient framework for optimizing expensive, noisy black-box functions $f: \mathcal{S} \to \mathbb{R}$ by leveraging a surrogate model and an acquisition function to balance exploration and exploitation~\cite{frazier2018tutorial}.
Multi-objective Bayesian Optimization (MOBO) extends BO to optimize multiple objectives $f_1, \ldots, f_M$,  aiming to efficiently approximate the Pareto frontier with respect to a user-specified reference point $r$. MOBO leverages specialized acquisition functions, such as Noisy Expected Hypervolume Improvement (NEHVI), to guide the search towards solutions that maximize the expected improvement in hypervolume~\cite{daulton2021qNEHVI,eriksson2021latencyaware}.

%% file: sections/3-architecture-search.tex
%% \vspace{-0.5em}

\begin{figure*}[!ht]
    \centering
    \includegraphics[width=\linewidth]{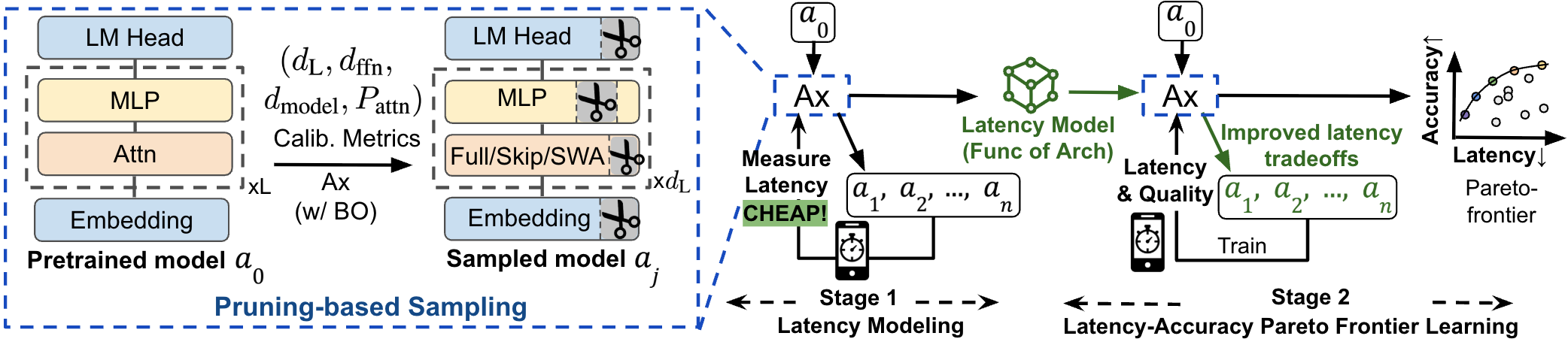}

    \caption{Overview of our two-stage OD-LLM design. We jointly search the architecture and attention pattern by pruning a pretrained model $a_0$ (Sec.~\ref{sec:search-space}) using Bayesian Optimization (BO) with Ax. In Stage~1, we sample pruned architectures and measure their latency on phone to cheaply learn a latency model. In Stage~2, leveraging the latency model, we efficiently search the space to generate the Accuracy-Latency Pareto-frontier (Sec.~\ref{sec:optimization-strategy}).
    }
    \label{fig:design process}

\end{figure*}

%% \vspace{-0.5em}
\section{Latency-Aware OD-LLM Design}
%% \vspace{-0.4em}
Our methodology addresses the challenge of designing OD-LLMs for resource-constrained devices.
We target the most common hardware platform and \textit{maximize model quality while minimizing TTFT}. 
Fig.~\ref{fig:design process} provides an overview.

%% \vspace{-0.5em}
\subsection{Latency-Aware Optimization}
\begin{figure}[h!]
    \centering
    \includegraphics[width=0.6\linewidth]{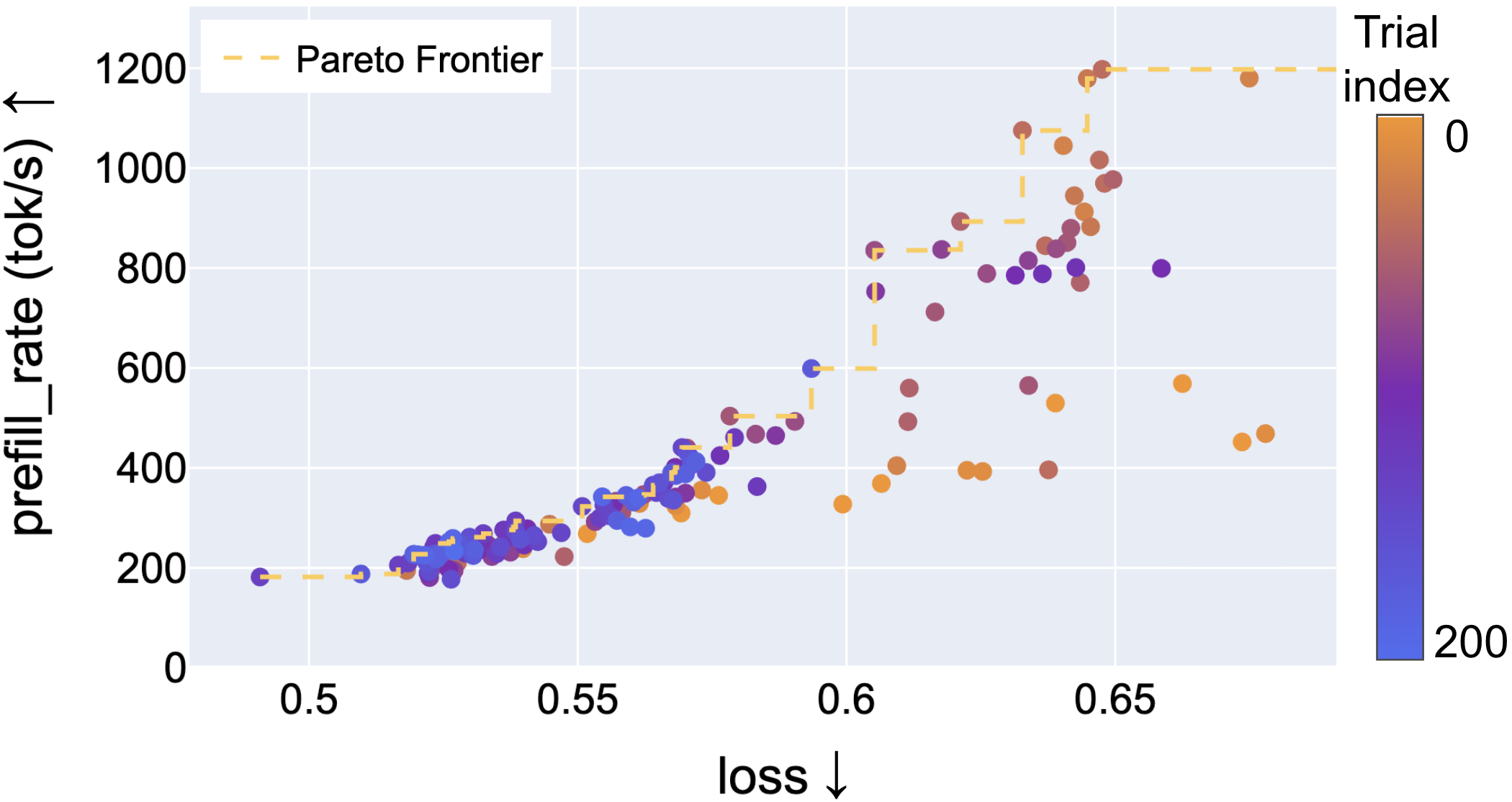}
    \vspace{-1em}
    \caption{A latency-loss Pareto-frontier}
    \label{fig:pareto-curve}
    % \vspace{-1em}
\end{figure}

%% \vspace{-0.4em}
We search candidate architectures $a \in \mathcal{A}$ for a configuration that maximizes model quality $Q(a)$ while meeting a constraint on our primary latency metric, TTFT, $T_\mathrm{prefill}(a) < \tau_\mathrm{prefill}$. We use loss as a proxy for $Q(a)$, based on the established findings that pretraining loss reliably predicts downstream quality~\cite{scaling-law,training-compute-optimal}. 
The threshold $\tau_\mathrm{prefill}$ is product-dependent and may change as the use-case evolves; because it is not known \textit{a priori}, we optimize both $Q$ and $T_\mathrm{prefill}$ jointly. 
Because improvements in latency often reduce quality, and vice versa, we target the Pareto frontier where no solution can improve one objective without worsening another. 
This frontier represents the set of optimal tradeoffs between the objectives as shown in Fig. \ref{fig:pareto-curve}. 
An optimal point can be selected from this frontier given any $\tau_\mathrm{prefill}$.

\subsection{Evaluating Model Configurations} %% \vspace{-0.4em}
To reduce the cost of the search relative to our target full training budget of $500$B tokens, we train each candidate architecture (obtained by pruning the base model) for only $2.6$B tokens. 
Empirically, we find that by $\sim 2.6$B tokens the relative ordering of architectures is already predictive of their ordering after the full $500$B-token run (Fig. \ref{fig:pruning_loss}). 
We refer to this early-but-predictive ordering as a \emph{stable ranking}. Due to resource constraints, we do \emph{not} sweep the optimizer hyperparameters.

% \begin{table}[h]
%   \centering
%   \caption{Correlation coefficients between real measured prefill / decode speed vs. parameter count / FLOPs.}
%   \label{tab:kendall_tau}
%   {\Large
%   \resizebox{\columnwidth}{!}{%
%     \begin{tabular}{|c|c|c|c|}\hline
%          Kendall tau / Spearman & prefill & decode \\\hline
%          \#params vs.\ latency per tok  &  0.40 / 0.55 &  0.40 / 0.68 \\\hline
%          FLOPs vs.\ latency per tok      &  0.46 / 0.63 & 0.55 / 0.73 \\\hline
%     \end{tabular}
%   }
%   }
% \end{table}

We emphasize that our focus is the on-device latency, and it is motivated by practical considerations. While total parameter counts and total FLOPs per token are important metrics, they do not fully determine the on-device latency. To be more concrete, in Tab.~\ref{tab:kendall_tau} we present the Kendall tau correlation coefficient between these quantities \emph{for our search space} across $100$ architectures exported with Executorch on a Samsung Galaxy S25 for $2$k sequence length. See Appendix~\ref{sec:appendix-efficiency-proxy} for visual representation of these correlations.

\begin{table}[ht]
%% \vspace{-0.3em}
\centering
\caption{Correlation coefficients between real measured prefill / decode speed vs. parameter count / FLOPs.}

\label{tab:kendall_tau}
\begin{adjustbox}{max width=0.6\columnwidth}
\begin{threeparttable}
%\resizebox{\columnwidth}{!}{
    \begin{tabular}{ccc}\hline
         Kendall tau correlation (\cite{kendall_tau})\tnote{*}  & prefill & decode \\\hline
         \#params vs.\ latency per tok  &  0.40  &  0.40 \\\hline
         FLOPs vs.\ latency per tok      &  0.46  & 0.55  \\\hline
    \end{tabular}
%}
\begin{tablenotes}
\item[*] \footnotesize {%Kendall Tau measures similarity between two rankings by quantifying the number of pairs that agree or disagree in their order. 
The closer to 1, the higher correlation.}
\end{tablenotes}
\end{threeparttable}
\end{adjustbox}
%% \vspace{-1em}
\end{table}

% For \textit{model efficiency}, we find low correlations between real measured prefill and decode latency per token versus number of model parameters and FLOPs, as shown in Table~\ref{tab:kendall_tau}. The results are based on 100 candidate architectures, exported using Executorch and benchmarked on a Samsung Galaxy S25 Ultra for 2k sequence length. More visualizations are shown in Appendix~\ref{sec:appendix-efficiency-proxy}. 

\begin{tcolorbox}[
  enhanced,
  colback=green!6!white,
  boxrule=0.8 pt, % Width of the border
  boxsep=0pt, % Space between text and the box frame
  left=2pt, % Left margin within the box
  right=2pt, % Right margin within the box
  top=2pt, % Top margin within the box
  bottom=2pt, % Bottom margin within the box
  drop fuzzy shadow=black!50
]
\textbf{Key Insight-1} Model parameter count and FLOPs are suboptimal proxies for latency; hardware-in-the-loop optimization is necessary for accurate on-device latency improvements.
\end{tcolorbox}

\subsection{Hybrid Architecture Search Space} %% \vspace{-0.5em}
\begin{figure}[!ht]
    \centering
\includegraphics[width=0.55\linewidth]{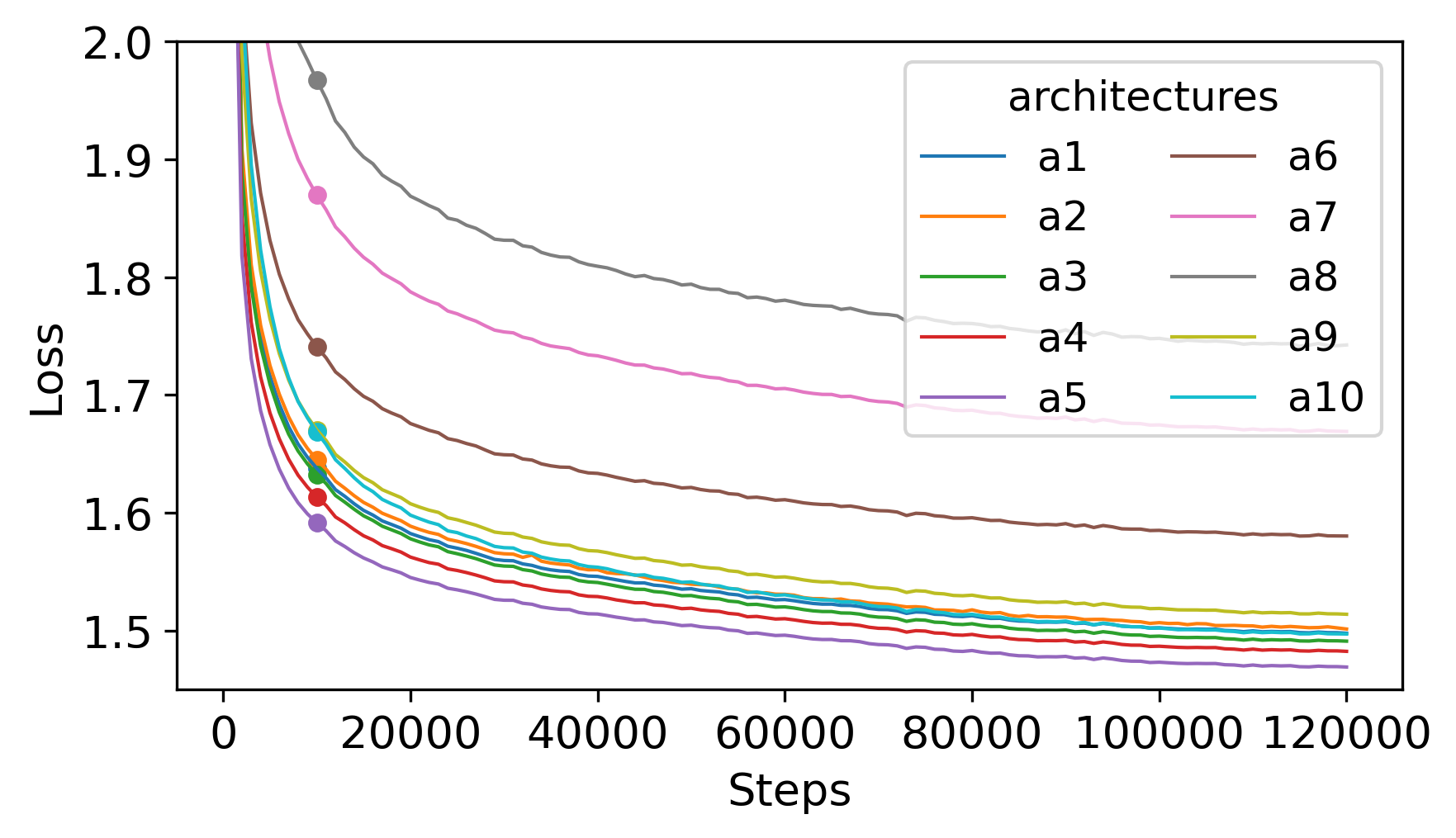}
\vspace{-1em}
    \caption{Pruned model loss evolution.}
    \label{fig:pruning_loss}

\end{figure}

\label{sec:search-space}
\noindent\textbf{Structured Pruning-based Search:} Instead of training candidates from scratch, we obtain candidates by pruning a larger pretrained model and inheriting its weights. Our approach is similar to weight-sharing approaches~\cite{fedorov2022udc, cai2019once, kusupati2022matryoshka} as the cost of training is amortized. However, we train candidates in isolation and are therefore not subject to the inductive bias of weight-sharing approaches, which assume that all candidates can be simultaneously trained in a nested fashion. We can think of a pruned pretrained model as a more data-efficient initialization for the architecture at hand. 
We observe empirically that the rank ordering of models trained from scratch aligns with that of pruned models (with the same architecture and inherited weights) under continued pretraining (CPT), with a Kendall tau correlation of $0.74$ across 20 candidates. 
Despite this alignment, CPT achieves a significantly lower loss and approaches final convergence values much faster than random initialization.
Furthermore, we need fewer tokens to get to a stable ranking of models; as shown in Fig. \ref{fig:pruning_loss}, the candidate ranking established at $10$k steps matches the ranking at $120$k steps. 

% % \vspace{-1em}

% \begin{table}[h]
% \centering
% \caption{Correlation between proxy evaluations and fully trained accuracy across xx candidate architectures. Pruning-based evaluation (which preserves learned weights) provides a stronger ranking signal than training from scratch under the same proxy evaluation and training budget.}
% \label{tab:proxy_correlation}
% {%\Large
% \resizebox{\columnwidth}{!}{
% \begin{tabular}{cccc}
% \hline
% Proxy evaluation method   & Spearman  & Kendall tau \\
% \hline
% Pruning from Large &  &  \\
% Train from Scratch &  &   \\
% \hline
% \end{tabular}
% }
% }
% \end{table}

Pruning can also be viewed as a natural method to efficiently explore the architecture search-space $\mathcal S$, consisting of the following parameters: number of transformer layers $d_\textrm{L}$, feed-forward network (FFN) hidden dimension $d_\textrm{ffn}$, residual stream dimension $d_\textrm{model}$, and efficient attention pattern $\mathbf{P}_\textrm{attn}$ (i.e., attention type for each transformer layer).
% Note that for each choice of $d_\textrm{L}, d_\textrm{ffn}, d_\textrm{model}$ and $\mathbf{P}_\textrm{attn}$ (meaning we choose an $s \in \mathcal S$), we still have many architectures $a\in \mathcal A$ that correspond to this choice. This ambiguity is resolved using activation-based metrics. 

We employ activation energy-based metrics, measured on a small calibration dataset to decide what to prune (similar to~\cite{fedorov2024llamaguard31bint4compact}). Specifically, the hidden size and MLP metrics quantify the average activation L2 norm magnitude (energy) across batch and sequence dimensions: $ \text{FFNMetric} = \frac{1}{N} \sum_{i=1}^{N} \|\mathbf{x}_i\| $ and $\text{ModelDimMetric} = \frac{1}{N} \sum_{i=1}^{N} \|\textrm{LN}(\mathbf{x}_i)\|$, while the layer metric captures the functional transformation energy via cosine similarity between input and output activations (as in \citet{gromov2024unreasonable}): $\text{LayerMetric} = 1 - \frac{1}{N} \sum_{i=1}^{N} \frac{\mathbf{x}_i \cdot \mathbf{x}_{i+1}}{\|\mathbf{x}_i\| \|\mathbf{x}_{i+1}\|}$, where the $\mathbf{x}_i$ is the input vector at position $i$, $N$ is the total number of positions (batch size $\times$ sequence length), and $\textrm{LN}$ is LayerNorm~\cite{ba2016layernormalization}. 
Guided by the the $\text{FFNMetric}$ and $\text{ModelDimMetric}$, we structurally prune the FFN hidden and residual stream dimensions for all decoder layers to the same target size $d_\textrm{ffn}$ and $d_\textrm{model}$ in block units (e.g., $128$). 
The setup of block size makes the resulting architecture compatible with group-wise quantization in post-training. 
We rank all decoder layers by $\text{LayerMetric}$ and remove those with the lowest contribution until $d_\textrm{L}$ layers remain. %reaching the target of $d_L$ remaining layers. 
After pruning, zeroed blocks and layers are removed, and the remaining architecture is concatenated into a compact and dense checkpoint.
%Using these metrics, we define a formal operator $\textrm{Prune}(\cdot) : \mathcal S \rightarrow \mathcal A$. 
%This operator takes $s=(L, d_\textrm{ffn}, d_\textrm{model})$ and produces a unique architecture $a\in \mathcal A$ that optimizes the activation metrics 
% \[
% \mathcal{A} = \left\{ a \mid a = \text{Prune}(L, d_\textrm{ffn}, d_\textrm{model}) \right\}\,.
% \]
 
\begin{tcolorbox}[
  enhanced,
  colback=green!6!white,
  boxrule=0.8 pt, % Width of the border
  boxsep=0pt, % Space between text and the box frame
  left=2pt, % Left margin within the box
  right=2pt, % Right margin within the box
  top=2pt, % Top margin within the box
  bottom=2pt, % Bottom margin within the box
  drop fuzzy shadow=black!50
]
\textbf{Key Insight-2} Pruning offers a more data-efficient approach to architecture search than training candidates from scratch. %, and better proxies candidate quality with a low training budget.
%and serves as a more reliable proxy for evaluating candidate architectures under a low training budget. 
\end{tcolorbox}

\noindent\textbf{Efficient Attention Pattern:}
\noindent Although many efficient attention mechanisms exist (Sec.~\ref{sec:related}), most prior work targets server-side GPU optimization. Our focus is on real-world, on-device deployment, requiring each candidate to be compiled and benchmarked within production-grade deployment stacks such as Executorch. This ensures our evaluation reflects practical constraints and operational realities of mobile and edge devices. Consequently, we restrict our search to efficient attention operations that are natively supported in Executorch, specifically (1) skip attention, which allows certain layers to bypass attention computation; (2) global attention and (3) sliding window attention (SWA)~\cite{child2019generatinglongsequencessparse}. Both global attention and SWA use grouped query and QK-Norm.  %Extending support for additional attention mechanisms in development stacks and expanding the search space is left for future work. 

Formally, the architecture space $\mathcal{A}$ consists of the model architecture and attention pattern $\mathbf{P}_\textrm{attn}$, where $\mathbf{P}_\textrm{attn} = <p_1, p_2, \ldots, p_\textrm{L}>$,  $i \in \textrm{L}$, $p_i$ specifies the attention type for the transformer layer $i$ in a model with $\textrm{L}$ layers. 
Using the importance metrics, we define a formal operator $\textrm{Prune}(\cdot) : \mathcal S \rightarrow \mathcal A$. 
This operator takes $s=(d_\textrm{L}, d_\textrm{ffn}, d_\textrm{model}, \mathbf{P}_\textrm{attn})$ and calibration metrics (FFNMetric, ModelDimMetric, LayerMetric) as input, and produces a unique architecture $a\in \mathcal A$ that optimizes the activation metrics. $\mathcal{A} = \left\{ a \mid a = \text{Prune}(\textrm{Calib. Metrics}, s) \right\}$.
Our search space (with $\sim 70B$ possible options) is shown in Tab.~\ref{tab:search_space}.

\begin{table}[!ht]
%% \vspace{-0.3em}
\centering
\caption{Search space $\mathcal{S}$.}
\label{tab:search_space}

\resizebox{0.4\columnwidth}{!}{%
\begin{tabular}{lc}
\hline
Parameters  &  Parameter Choices \\
\hline
$d_\textrm{L}$ &  $\{10,11,12,13,14,15,16\}$  \\
$d_\textrm{ffn}$ & $\{2048, 2304, 2560, \ldots 8192\}$    \\
$d_\textrm{model}$ & $\{1024, 1152, 1280 \ldots 2048\} $   \\
$p_i$ & \{full\_attn, SWA, skip\_attn\} \\
\hline
\end{tabular}
}

\end{table}

\subsection{Optimization Strategy}\label{sec:optimization-strategy}
We leverage Bayesian optimization (BO) with the Ax platform~\cite{olson2025ax} to optimize for the latency–quality Pareto frontier by searching over $\mathcal{S}$. 
% Its ability to efficiently optimize multiple simultaneous objectives in mixed discrete-and-continuous search spaces is essential for navigating critical tradeoffs between model cost and quality ~\cite{eriksson2021latencyaware}.
We use a \textit{two-stage BO} approach to leverage the fact that latency is nearly instantaneous while model quality requires substantial training resources. 
In the first stage, we use Sobol quasi-random sampling~\cite{SOBOL196786} to densely cover the latency landscape of $\mathcal{S}$, followed by BO. 
Since latency evaluation is significantly cheaper than quality evaluation, we collect ${\sim}800$ on-device latency measurements to build a high-quality Gaussian Process surrogate, achieving a cross-validation $R^2 = 0.97$. This allows the second stage to rely on predicted latency, focusing expensive quality evaluations on architectures in latency-favorable regions. 
%In the first stage, we densely sample the latency landscape of $\mathcal{S}$ to build a high-quality Gaussian Process surrogate model. 
The second stage optimizes both objectives, accelerating the multi-objective search by focusing expensive model-quality evaluations on regions more likely to exhibit favorable latency tradeoffs. 
While BO methods such as HVKG~\cite{pmlr-v202-daulton23a} support objectives with different evaluation costs, our setting assumes latency is essentially free to evaluate. 
% Therefore, we use $q\mathrm{NEHVI}$\cite{daulton2021qNEHVI}, maximizes the expected improvement in hypervolume $\mathcal{H}(\mathcal{P})$, where $\mathcal{P}$ is the set of observed objective vectors. The hypervolume measures the multi-dimensional area dominated by the Pareto frontier relative to a reference point $r$. Specifically, $q\mathrm{NEHVI}$ is defined as
% \begin{equation}
%     \alpha_{q\mathrm{NEHVI}}(s)=\mathbb{E}\left(\mathcal{H}(\mathcal{P}\cup \mathcal{GP}(s)) - \mathcal{H}(\mathcal{P})\right)
% \end{equation}
% where the expectation is taken over the joint distribution of the objective values at $s$. 
We leverage the NEHVI acquisition function and set the reference point $r$ to a loss of $0.6$ and a TTFT of $4$ seconds, focusing on configurations that outperform a hand-tuned baseline model. The loss threshold of $0.6$ is a heuristic calibrated by training a small number of candidates and validating against downstream tasks, and based on our own rank-ordering stability analysis (Sec~\ref{sec:search-space}).

% (Table ~\ref{tab:feasibility_comparison}).

% \begin{table}[h]
% \centering
% \caption{Feasibility rates (valid vs. total sampled candidates) with and without latency-aware warm-starting.}
% \label{tab:feasibility_comparison}
% \resizebox{0.8\columnwidth}{!}{%
% \begin{tabular}{cc}
% \hline
%   &  Feasibility rates \\
% \hline
% With warm-starting &  47.73\%  \\
% Without warm-starting &     \\
% \hline
% \end{tabular}
% }
% \end{table}

% \vspace{-1.2em}
\subsection{OD-LLM Tuning Principles}\label{sec:efficiency-principles}
\begin{figure}[h]
    \centering
    \includegraphics[width=0.6\linewidth]{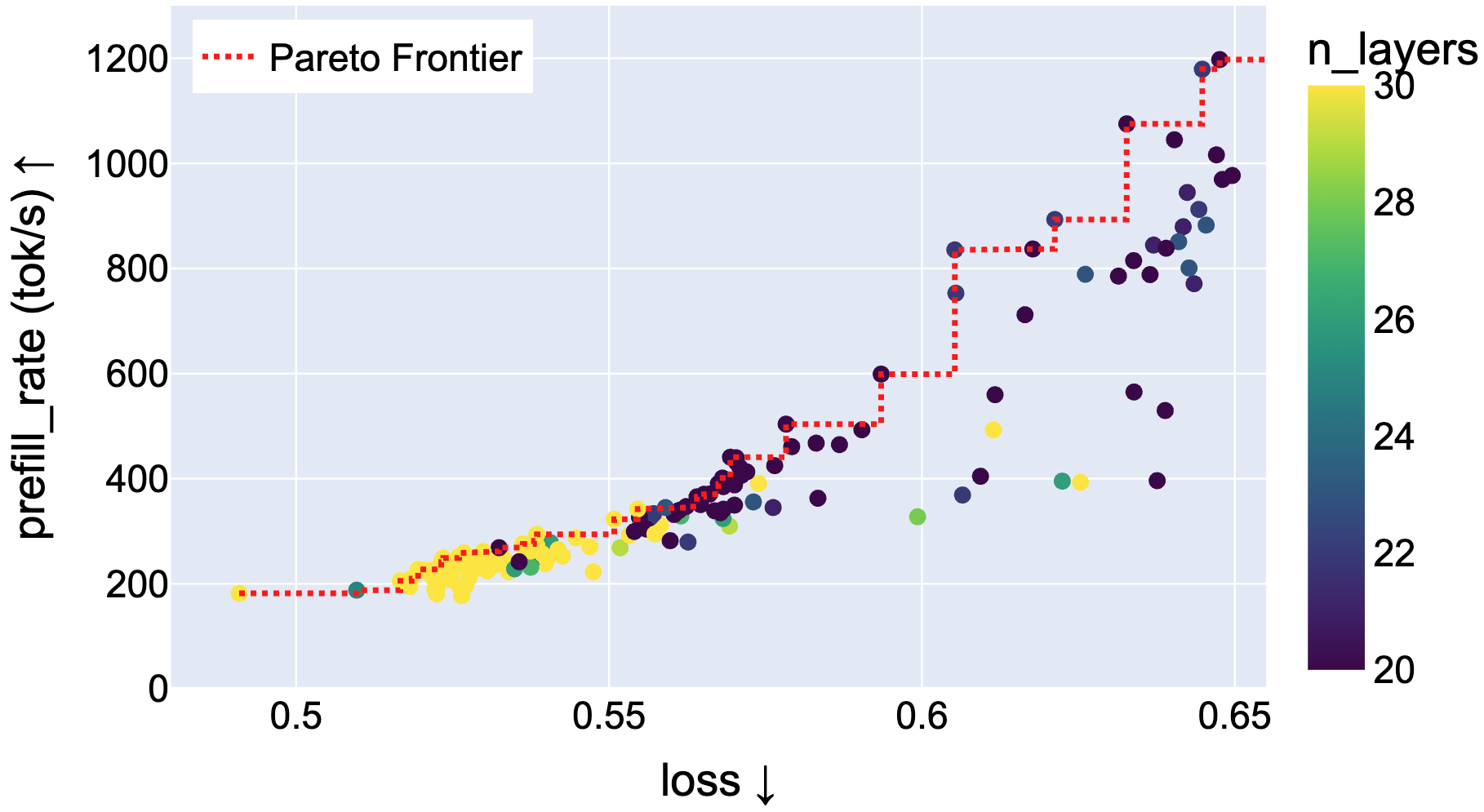}
    % \vspace{-1.5em}
    \caption{A Pareto curve with different model depths.}
    \label{fig:pareto-curve_n_layer}
    %% \vspace{-1.5em}
\end{figure}

%% \vspace{-0.5em}
By analyzing the optimal candidates along the Pareto-frontier, we distill the following efficiency principles for latency-aware OD-LLM design:

\noindent (1) \textbf{Model architecture}: We find that (at a fixed parameter count) deeper models generally have lower loss and higher latency, while shallow-and-wide models have higher loss and lower latency. Yet, at sufficiently low latency it is preferable to switch to shallow models. As shown in Fig.~\ref{fig:pareto-curve_n_layer}, on the Pareto-frontier curve, the $30$-layer models (yellow dots) achieve the best model quality with the slowest prefill speed, while the shallower architectures (dark blue dots) offer a better balance on accuracy–latency trade-off. This aligns with observations in prior work~\cite{fu2025nemotronflash}.

\begin{tcolorbox}[
  enhanced,
  colback=blue!6!white,
  boxrule=0.8 pt, % Width of the border
  boxsep=0pt, % Space between text and the box frame
  left=2pt, % Left margin within the box
  right=2pt, % Right margin within the box
  top=2pt, % Top margin within the box
  bottom=2pt, % Bottom margin within the box
  drop fuzzy shadow=black!50
]
\textbf{Efficiency Principle-1} %Deeper and thinner architectures tend to yield higher model quality, while shallower and wider architectures achieve better latency. 
For on-device deployment, shallow-and-wide models better balance accuracy and latency than deeper models.
\end{tcolorbox}

\noindent (2) \textbf{Attention pattern}: We find, surprisingly, that our latency-guided search consistently favors skip attention over SWA, indicating that skipping a module provides a better latency-quality trade-off than SWA. 
% The selection of attention patterns is affected by the real hardware characteristics and memory constraints (e.g., prefill chunk size) considered during the search process.
We provide a high-level explanation for the inefficiency of SWA in our settings in Appendix~\ref{inefficient-SWA}. 
%A high level explanation for the inefficiency of SWA in our settings is as follows. We find that a prefill chunk size of 1024 produces the lowest time-to-first-token (TTFT), making the best use of parallelism across tokens in the context window. At the same time, Executorch constrains the SWA to be greater than or equal to the prefill chunk size, such that the sliding window is lower-bounded by $1024$. Therefore, when prefilling $2$k tokens, only the second chunk benefits from the SWA. At the same time, the ring-buffer implementation of SWA in Executorch requires computation of the entire attention matrix, as opposed to just the lower triangular portion as done during standard attention computation. As a result, in the $2$k prefill, $1024$ chunk size setting, we find that the benefits of SWA are outweighed by the drawback of the slower attention calculation and the model overall achieves worse TTFT with SWA.
Moreover, we observe that skipping too many attention layers consecutively (empirically, more than $3$) degrades model quality under a fixed latency constraint and can even impair performance on harder generative tasks, as shown in Tab.~\ref{tab:consecutive_skip}. %Intuitively, because attention is essential for mixing information across tokens, excessive skipping limits updates to position-wise transforms (e.g., MLPs), thereby hindering the integration of global context. In contrast, inserting global-attention blocks periodically preserves the model’s ability to propagate long-range dependencies. 
Consequently, we find the optimal attention pattern interleaves skip attention and global attention blocks. Therefore, we add a constraint in the final search: don't include $3$ or more consecutive efficient attention types (either SWA or skip attention).

\begin{table}[!ht]
\centering
\caption{Candidates with identical latency / identical architecture and skip attention counts can differ in performance due to the consecutive attention patterns.}
%Candidates with the same latency performance can behave differently due to consecutive attention skipping. Both candidates have the same $d_\textrm{L}$, $d_\textrm{ffn}$, $d_\textrm{model}$ and number of skip attentions, but only different attention patterns.
%}
\label{tab:consecutive_skip}

\resizebox{0.52\columnwidth}{!}{
\begin{tabular}{cccc}
\hline
  &  TQA & NQ & WinoG \\
\hline
With >3 consecutive skip attention &  8.8\% & 2.5\% & 59.7\%   \\
Without >3 consecutive skip attention & 33.2\% & 10.0\% & 61.2\%   \\
\hline
\end{tabular}}
%% \vspace{-1em}
\end{table}

\begin{tcolorbox}[
  enhanced,
  colback=blue!6!white,
  boxrule=0.8 pt, % Width of the border
  boxsep=0pt, % Space between text and the box frame
  left=2pt, % Left margin within the box
  right=2pt, % Right margin within the box
  top=2pt, % Top margin within the box
  bottom=2pt, % Bottom margin within the box
  drop fuzzy shadow=black!50
]
\textbf{Efficiency Principle-2} 
Skip attention is more efficient than SWA; optimal patterns interleave skip attention and global attention to balance long-range modeling and latency.
%Skip attention offers greater efficiency compared to SWA. The optimal attention pattern interleaves skip attention with global attention blocks to balance long-range dependency modeling with latency.
\end{tcolorbox}

%% file: sections/4-results.tex
\section{\sys{}: A New Family of Fast On-Device LLMs}
%% \vspace{-0.5em}

\subsection{Implementation}

Our pruning-based architecture search is implemented as follows.
\textbf{(1) Starting checkpoint}: As Sec.~\ref{sec:efficiency-principles} highlights the benefits of shallow architectures, we select MobileLLM-Pro-Shallow-1.8B (Tab.~\ref{tab:backbone}) as the starting point -- a shallow variant of MobileLLM-Pro~\cite{mobilellmpro} trained with the same recipe and data. 
We use the same pretraining data, instruction fine-tuning (IFT) data and tokenizer ($202$k vocabulary size) as MobileLLM-Pro~\cite{mobilellmpro} throughout our experiment. 
\textbf{(2) Calibration}: We calibrate the activation-based importance metrics (Sec.~\ref{sec:search-space}) on a $600$M-token (${\sim}0.1\%$ of the pretraining tokens) calibration set, following common practice for structured pruning~\cite{fedorov2024llamaguard31bint4compact,gromov2024unreasonable}. 
\textbf{(3) Sampling and pruning}: Ax samples $8$ candidates per iteration, determined by available GPU resources for parallel evaluation within each BO round. We prune the  initial model based on these configurations, applying efficient attention patterns and inheriting weights to create hybrid small, dense models. 
\textbf{(4) Candidate evaluation}: Candidate architectures are trained for $2.6$B tokens to ensure stable rank-ordering of architectures (as discussed in Sec. ~\ref{sec:search-space}) and the loss is used to measure model quality. We then export models with ExecuTorch (v1.1.0) and measure TTFT on a Samsung Galaxy S25 at a $2$k sequence length. 
\textbf{(5) Candidate selection and final training}:  We run $200$ trials to maximize coverage of the latency--quality tradeoff space within our compute budget. After $200$ trials, we select Pareto-optimal candidates that satisfy our specific latency constraints. Then we CPT the best candidates for $500$B tokens using knowledge distillation with MobileLLM-Pro-Shallow-1.8B as teacher. 

Our approach only requires lightweight CPT rather than training from scratch: we use $\sim35\%$ of the tokens used to pretrain MobileLLM-Pro. Finally, the models are IFTed for $800$B tokens to prepare them for downstream tasks. During CPT, we set the sequence length to 2048, and the SWA window size to 256. During IFT, we set the sequence length to 8192, enabling the model to generalize to longer-sequence downstream tasks. Note that the total search cost is $200 \times 2.6\text{B} = 520\text{B}$ tokens across all candidates. Only 3 final Pareto-optimal candidates undergo full training: 500B tokens of CPT and 800B tokens of IFT each. This makes our method practical for OD-LLM development, compared to MobileLLM-Pro (1.6T)~\cite{mobilellmpro} and LFM2 (10-12T)~\cite{LFM2}.

\begin{table}[!ht]
%% \vspace{-0.3em}
\centering
\caption{MobileLLM-Pro variants and \sys{} model architectures.}
\label{tab:backbone}
\begin{adjustbox}{max width=0.6\columnwidth}
\begin{threeparttable}
%\resizebox{\columnwidth}{!}{
\begin{tabular}{lccccc}
\toprule
\textbf{Model} & Layers & $d_{\text{model}}$ & $d_{\text{FFN}}$ & H/KV/$H_{\text{size}}$ &  \makecell{\#Attn\\blocks} \\
\midrule
MobileLLM-Pro-1B\tnote{*} & 30 & 1280 & 6144 & 20/4/64 & 16 \\
MobileLLM-Pro-Shallow-1.8B & 16 & 2048 & 8192 & 32/8/64 & 16 \\
\sys{}-350M\tnote{*} & 12 & 1024 & 4096 & 32/8/64 & 7\tnote{$\dagger$} \\
& &  \multicolumn{4}{r}{\textit{ full\_attn\_idxs: [0,1,3,6,7,9,11]\tnote{$\dagger$}}} \\
\sys{}-650M\tnote{*} & 13 & 1280 & 6144 & 32/8/64 & 8\tnote{$\dagger$} \\
& &  \multicolumn{4}{r}{\textit{ full\_attn\_idxs: [0,2,3,5,7,8,9,10]\tnote{$\dagger$}}} \\
\sys{}-1.4B\tnote{*} & 16 & 2048 & 8192 & 32/8/64 & 16 \\
\bottomrule
\end{tabular}
%}
\begin{tablenotes}
\item[$\dagger$] The remaining layers skip attention.
\item[*] Shared weights between input embeddings and output projection.

\end{tablenotes}
\end{threeparttable}
\end{adjustbox}
%% \vspace{-1em}
\end{table}

%% \vspace{-0.3em}
\subsection{Experimental Results}\label{sec:experimental-results}
%% \vspace{-0.5em}

\begin{table*}[!ht]
\centering
\caption{Comparison of model quality and efficiency performance across various on-device scale models}
\vspace{-0.5em}
\label{tab:model_performance}
\Large{
\resizebox{\textwidth}{!}{
\begin{tabular}{l|cccccccccc|ccc|ccc}
\toprule
Model (Parameter Count)
& HellaSwag & BoolQ & PIQA & SocialIQA & TQA & NQ & ARC-c & ARC-e & WinoG & Avg $\uparrow$
& \multicolumn{3}{c|}{Prefill TTFT(s) $\downarrow$}
& \multicolumn{3}{c}{Decode rate (tok/s) $\uparrow$} \\
\addlinespace[-0.2em]
% Dataset citations (separate header line)
& \scriptsize\cite{hellaswag}
& \scriptsize\cite{boolq}
& \scriptsize\cite{piqa}
& \scriptsize\cite{siqa}
& \scriptsize\cite{triviaqa}
& \scriptsize\cite{nq}
& \scriptsize\cite{arcc} % ARC-c
& \scriptsize\cite{arcc} % ARC-e (use a different key if you have one)
& \scriptsize\cite{winogrande}
&
& \multicolumn{3}{c|}{}
& \multicolumn{3}{c}{} \\
\cmidrule(lr){2-11}\cmidrule(lr){12-14}\cmidrule(lr){15-17}
& & & & & & & & & &
& 1k & 2k & 4k
& 1k & 2k & 4k \\
\midrule

Gemma3 270M \newline {\scriptsize\cite{gemmateam2025gemma3technicalreport}}
& 41.38 & 58.17 & 68.34 & 39.71 & 15.4 & 4.04 & 29.0 & 57.32 & 53.59 & 40.77
& 0.99 & 3.13 & 5.64
& 123.71 & 82.85 & 52.40 \\
LFM2 350M \newline {\scriptsize\cite{LFM2}}
& 49.00 & \textbf{64.37} & 69.48 & 35.01 & 14.97 & 4.96 & \textbf{44.54} & \textbf{66.04} & 55.96 & 44.92
& \textbf{0.84} & \textbf{2.18} & \textbf{4.52}
& 147.54 & 96.90 & 63.60 \\
\rowcolor{lightgreen}
\sys{} 350M
& \textbf{49.16} & 62.39 & \textbf{70.08} & \textbf{43.50} & \textbf{19.84} & \textbf{5.90} & 38.89 & 63.26 & \textbf{56.09} & \textbf{45.46}
& 0.91 & 2.78 & 5.33
& \textbf{165.56} & \textbf{112.58} & \textbf{95.55} \\
\midrule

Qwen3 0.6B \newline {\scriptsize\cite{yang2025qwen3technicalreport}}
& 53.80 & 69.39 & 69.86 & 43.14 & 2.41 & 4.32 & 38.57 & 58.08 & 58.88 & 44.27
& 4.63 & 11.59 & 25.40
& 44.56 & 28.67 & 18.48 \\
LFM2 700M \newline {\scriptsize\cite{LFM2}}
& 45.01 & \textbf{71.38} & 71.11 & 37.41 & 22.50 & 6.90 & \textbf{49.40} & \textbf{74.60} & 58.40 & 48.52
& 2.05 & 6.01 & \textbf{8.24}
& 86.30 & 53.57 & 44.36 \\
\rowcolor{lightgreen}
\sys{} 650M
& \textbf{54.60} & 64.77 & \textbf{71.82} & \textbf{45.45} & \textbf{24.49} & \textbf{7.56} & 42.58 & 66.55 & \textbf{59.35} & \textbf{48.57}
& \textbf{1.62} & \textbf{3.34} & 8.48
& \textbf{96.64} & \textbf{85.35} & \textbf{60.74} \\
\midrule

Nemotron-Flash 1B \newline {\scriptsize\cite{fu2025nemotronflash}}
& 45.80 & -- & 75.41 & -- & -- & -- & 41.47 & 74.83 & 59.67 & --
& -- & -- & --
& -- & -- & -- \\
Gemma3 1B \newline {\scriptsize\cite{gemmateam2025gemma3technicalreport}}
& 62.30 & 63.20 & 73.80 & \textbf{48.90} & \textbf{39.80} & 9.48 & 38.40 & 73.00 & 58.20 & 51.79
& 3.55 & 9.29 & 18.86
& 58.58 & \textbf{43.11} & \textbf{36.29} \\
Llama3.2 1B \newline {\scriptsize\cite{grattafiori2024llama3herdmodels}}
& 65.69 & 62.51 & 75.14 & 45.60 & 23.81 & 5.48 & 38.28 & 63.47 & 61.09 & 49.01
& 4.51 & 15.09 & 26.95
& 39.01 & 18.13 & 14.34 \\
LFM2 1.2B \newline {\scriptsize\cite{LFM2}}
& 45.26 & 66.09 & 74.27 & 37.72 & 33.50 & 8.60 & \textbf{52.20} & \textbf{77.90} & 58.80 & 50.48
& 3.46 & \textbf{8.41} & 16.88
& \textbf{61.14} & 42.15 & 29.14 \\
\rowcolor{lightgreen}
\sys{} 1.4B
& \textbf{66.87} & \textbf{71.07} & \textbf{75.52} & 47.34 & 36.06 & \textbf{11.83} & 50.56 & 72.26 & \textbf{64.01} & \textbf{55.06}
& \textbf{3.40} & 9.08 & \textbf{16.50}
& 60.52 & 42.65 & 34.01 \\
\midrule \multicolumn{17}{l}{\textit{Base model (before pruning)}} \\ MobileLLM-Pro-Shallow-1.8B & 67.74 & 72.63 & 76.82 & 47.34 & 39.00 & 12.41 & 51.33 & 74.12 & 64.48 & 56.20 & 3.82 & 9.20 & 19.93 & 46.52 & 35.88 & 25.69 \\
\bottomrule
\end{tabular}
}
}
\end{table*}

\begin{table}[t!]
\centering
\caption{Comparison of downstream task results}
\label{tab:ift-results}
\vspace{-0.5em}
\resizebox{0.6\columnwidth}{!}{
\begin{tabular}{cccccc}
\hline
  &  MMLU & MBPP &  HumanEval & Open Rewrite & TLDR9+ \\
\hline
Gemma3-1B  & 29.90&35.20& 41.50& -- & --\\
Llama3.2-1B & \textbf{49.30}&\textbf{39.60}& 37.80& 41.60 & 16.80 \\
\sys{} 650M  &35.37&33.00 & \textbf{45.12}& \textbf{46.84} & 14.93\\
\sys{} 1.4B  &47.89 & 35.60& \textbf{46.34} & 40.10 & \textbf{16.89}\\
\hline
\end{tabular}}

\end{table}

\begin{table}[t!]
\centering
\caption{MobileLLM-Flash maintains competitive or superior TTFT across devices.}
\label{tab:cross_device}
\vspace{-0.5em}
\resizebox{0.3\columnwidth}{!}{
\begin{tabular}{lcc} 
\toprule TTFT 1k (s) $\downarrow$ & S25 & iPhone 17 \\ 
\midrule LFM2 350M & \textbf{0.84} & \textbf{1.47} \\ 
\sys{} 350M & 0.91 & 1.55 \\ 
\midrule LFM2 700M & 2.05 & 3.40 \\ 
\sys{} 650M & \textbf{1.62} & \textbf{2.81} \\ 
\midrule LFM2 1.2B & 3.46 & 5.51 \\ 
\sys{} 1.4B & \textbf{3.40} & \textbf{4.96} \\ 
\bottomrule \end{tabular} 
}
\vspace{-1em}
\end{table}

\subsubsection{Model Quality}
We evaluate our pre-trained models across reasoning, retrieval, and knowledge-intensive tasks: HellaSwag~\cite{hellaswag}, BoolQ~\cite{boolq}, PIQA~\cite{piqa}, SIQA~\cite{siqa}, \textit{WinoG}rande~\cite{winogrande}, \textit{ARC} \textit{E}asy~\cite{arcc} in 0-shot settings, \textit{T}rivia\textit{QA}~\cite{triviaqa} and \textit{N}at\textit{Q}~\cite{nq} with 5 shots and \textit{ARC} \textit{C}hallenge with 25 shots. Using lm-eval-harness~\cite{lm-eval-harness}, we report exact-match rate for TriviaQA/NQ and character-level accuracy for other tasks. We compare model accuracy and efficiency with OD-LLM baselines LFM2-$350$M/$700$M/$1.2$B~\cite{LFM2}, Nemotron-FLash-$1$B~\cite{fu2025nemotronflash}, Qwen3 $0.6$B~\cite{yang2025qwen3technicalreport}, Llama3.2-$1$B~\cite{grattafiori2024llama3herdmodels} and Gemma3-$270$M/$1$B~\cite{gemmateam2025gemma3technicalreport}, as shown in Tab.~\ref{tab:model_performance}. \sys{} achieves the highest average scores across all size regimes, establishing it as the most capable on-device model. For reference, Tab.~\ref{tab:model_performance} includes the unpruned MobileLLM-Pro-Shallow-1.8B base model. \sys{}-1.4B trades a modest $1.1\%$ average accuracy for up to $1.2\times/1.3\times$ faster prefill/decode, illustrating the favorable latency-quality tradeoff identified by our search.
%The results demonstrate that our model consistently achieves superior or on-par performance compared to baselines, with all the average scores outperforming all the baselines across various model size regimes, establishing \sys{} as the most capable generalist language model at the on-device scale.

We evaluate our IFTed models across knowledge (MMLU), coding (MBPP, HumanEval), rewriting (OpenRewrite), and summarization (TLDR9+) tasks (Tab.~\ref{tab:ift-results}). All tasks are formatted as user-assistant conversations and evaluated on the final response. The results demonstrate that \sys{} achieves superior or comparable performance, making it as the leading on-device instruction-tuned model for assistant applications.

\vspace{-0.5em}
\subsubsection{Model Efficiency}
\vspace{-0.5em}

For latency evaluation, we export models using ExecuTorch (v1.1.0) and optimize them for mobile CPUs via the XNNPACK backend. All models are quantized to 4-bit weights (group size 32) and 8-bit dynamic activations, with a quantized KV cache. Nemotron-Flash-1B is excluded because its custom JetBlock operators are not supported by ExecuTorch. We conduct latency benchmarking on the Samsung Galaxy S25, a flagship Android phone powered by the Snapdragon 8 Elite chipset with an octa-core CPU and 12 GB of memory. We evaluated models at context lengths of 1k, 2k, and 4k using 4 CPU threads. Reported metrics are averaged over three runs following a  warmup. The results demonstrate that \sys{} yields \bm{$1.8\times / 1.6\times$} faster prefill / decode speeds than LFM2, establishing \sys{} as the fastest OD-LLM at this scale.

%IGOR: Mention vocabulary size; maybe: larger vocab == smaller sequence length
Our models use a 202k vocabulary size. While this increases the embedding parameters, it enhances information density per token. By representing common words more efficiently, the model requires fewer tokens to encode the same semantic content compared to models with smaller vocabulary sizes (e.g., 32k), leading to cheaper inference. Besides, this vocabulary is shared with Llama4-Scout~\cite{meta2025llama4}, enabling effective knowledge distillation that benefits both our shallow and deep backbones~\cite{mobilellmpro}.

\sys{} uses only standard operators natively supported by ExecuTorch and thus deploys on any compatible backend — including Apple devices (CoreML dispatches to CPU, GPU, Apple Neural Engine) — without specialized kernels. Retargeting the search to a new platform requires only substituting the benchmark device; the methodology itself is hardware-agnostic. We note that Pareto-optimal architectures for one accelerator class (e.g., CPU) are not necessarily optimal for another (e.g., Apple Neural Engine) due to differing execution characteristics. Nevertheless, we observe that latency rankings tend to transfer well across mobile CPUs: architectures searched on a Samsung Galaxy S25 preserve their relative advantage on an iPhone~17 (Tab.~\ref{tab:cross_device}).

% \subsection{Pruning-Based vs. From Scratch Training}\label{sec:results_pruning_vs_scratch}

%% file: paper.bib
@inproceedings{
yang2025zebrallama,
title={Zebra-Llama: Towards Extremely Efficient Hybrid Models},
author={Mingyu Yang and Mehdi Rezagholizadeh and Guihong Li and Vikram Appia and Emad Barsoum},
booktitle={The Thirty-ninth Annual Conference on Neural Information Processing Systems},
year={2025},
url={https://openreview.net/forum?id=l42UGsdrNn}
}

@article{kim2026seconds,
  title={From seconds to sentiments: differential effects of chatbot response latency on customer evaluations},
  author={Kim, Kaeun and Shams, Ghazal and Kim, Kawon},
  journal={International Journal of Human--Computer Interaction},
  volume={42},
  number={1},
  pages={597--612},
  year={2026},
  publisher={Taylor \& Francis}
}

@article{kusupati2022matryoshka,
  title={Matryoshka representation learning},
  author={Kusupati, Aditya and Bhatt, Gantavya and Rege, Aniket and Wallingford, Matthew and Sinha, Aditya and Ramanujan, Vivek and Howard-Snyder, William and Chen, Kaifeng and Kakade, Sham and Jain, Prateek and others},
  journal={Advances in Neural Information Processing Systems},
  volume={35},
  pages={30233--30249},
  year={2022}
}

@article{cai2019once,
  title={Once-for-all: Train one network and specialize it for efficient deployment},
  author={Cai, Han and Gan, Chuang and Wang, Tianzhe and Zhang, Zhekai and Han, Song},
  journal={arXiv preprint arXiv:1908.09791},
  year={2019}
}

@article{fedorov2022udc,
  title={UDC: Unified DNAS for compressible TinyML models for neural processing units},
  author={Fedorov, Igor and Matas, Ramon and Tann, Hokchhay and Zhou, Chuteng and Mattina, Matthew and Whatmough, Paul},
  journal={Advances in Neural Information Processing Systems},
  volume={35},
  pages={18456--18471},
  year={2022}
}

@book{nielsen1994usability,
  title={Usability engineering},
  author={Nielsen, Jakob},
  year={1994},
  publisher={Morgan Kaufmann}
}

@inproceedings{heinecke2017understanding,
  title={Understanding the performance of small convolution operations for CNN on intel architecture},
  author={Heinecke, Alexander and Georganas, Evangelos and Banerjee, Kunal and Kalamkar, Dhiraj and Sundaram, Narayanan and Venkat, Anand and Henry, Greg and Pabst, Hans},
  booktitle={Poster in the International Conference for High Performance Computing, Networking, Storage, and Analysis},
  year={2017}
}

@misc{fedorov2024llamaguard31bint4compact,
      title={Llama Guard 3-1B-INT4: Compact and Efficient Safeguard for Human-AI Conversations}, 
      author={Igor Fedorov and Kate Plawiak and Lemeng Wu and Tarek Elgamal and Naveen Suda and Eric Smith and Hongyuan Zhan and Jianfeng Chi and Yuriy Hulovatyy and Kimish Patel and Zechun Liu and Changsheng Zhao and Yangyang Shi and Tijmen Blankevoort and Mahesh Pasupuleti and Bilge Soran and Zacharie Delpierre Coudert and Rachad Alao and Raghuraman Krishnamoorthi and Vikas Chandra},
      year={2024},
      eprint={2411.17713},
      archivePrefix={arXiv},
      primaryClass={cs.DC},
      url={https://arxiv.org/abs/2411.17713}, 
}

@misc{LFM2,
      title={LFM2 Technical Report}, 
      author={Alexander Amini and Anna Banaszak and Harold Benoit and Arthur Böök and Tarek Dakhran and Song Duong and Alfred Eng and Fernando Fernandes and Marc Härkönen and Anne Harrington and Ramin Hasani and Saniya Karwa and Yuri Khrustalev and Maxime Labonne and Mathias Lechner and Valentine Lechner and Simon Lee and Zetian Li and Noel Loo and Jacob Marks and Edoardo Mosca and Samuel J. Paech and Paul Pak and Rom N. Parnichkun and Alex Quach and Ryan Rogers and Daniela Rus and Nayan Saxena and Bettina Schlager and Tim Seyde and Jimmy T. H. Smith and Aditya Tadimeti and Neehal Tumma},
      year={2025},
      eprint={2511.23404},
      archivePrefix={arXiv},
      primaryClass={cs.LG},
      url={https://arxiv.org/abs/2511.23404}, 
}

@article{gromov2024unreasonable,
  title={The unreasonable ineffectiveness of the deeper layers},
  author={Gromov, Andrey and Tirumala, Kushal and Shapourian, Hassan and Glorioso, Paolo and Roberts, Daniel A},
  journal={arXiv preprint arXiv:2403.17887},
  year={2024}
}

@article{cowsik2025towards,
  title={Towards Distributed Neural Architectures},
  author={Cowsik, Aditya and He, Tianyu and Gromov, Andrey},
  journal={arXiv preprint arXiv:2506.22389},
  year={2025}
}

@inproceedings{
fu2025nemotronflash,
title={Nemotron-Flash: Towards Latency-Optimal Hybrid Small Language Models},
author={Yonggan Fu and Xin Dong and Shizhe Diao and Matthijs Van keirsbilck and Hanrong Ye and Wonmin Byeon and Yashaswi Karnati and Lucas Liebenwein and Maksim Khadkevich and Alexander Keller and Jan Kautz and Yingyan Celine Lin and Pavlo Molchanov},
booktitle={The Thirty-ninth Annual Conference on Neural Information Processing Systems},
year={2025},
url={https://openreview.net/forum?id=KTDAbnFsQj}
}

@inproceedings{
gu2025jetnemotron,
title={Jet-Nemotron: Efficient Language Model with Post Neural Architecture Search},
author={Yuxian Gu and Qinghao Hu and Haocheng Xi and Junyu Chen and Shang Yang and Song Han and Han Cai},
booktitle={The Thirty-ninth Annual Conference on Neural Information Processing Systems},
year={2025},
url={https://openreview.net/forum?id=WZQXaTNYEB}
}

@misc{mobilellmpro,
      title={MobileLLM-Pro Technical Report}, 
      author={Patrick Huber and Ernie Chang and Wei Wen and Igor Fedorov and Tarek Elgamal and Hanxian Huang and Naveen Suda and Chinnadhurai Sankar and Vish Vogeti and Yanghan Wang and Alex Gladkov and Kai Sheng Tai and Abdelrahman Elogeel and Tarek Hefny and Vikas Chandra and Ahmed Aly and Anuj Kumar and Raghuraman Krishnamoorthi and Adithya Sagar},
      year={2025},
      eprint={2511.06719},
      archivePrefix={arXiv},
      primaryClass={cs.LG},
      url={https://arxiv.org/abs/2511.06719}, 
}

@article{frazier2018tutorial,
  title={A tutorial on Bayesian optimization},
  author={Frazier, Peter I},
  journal={arXiv preprint arXiv:1807.02811},
  year={2018}
}

@article{liu2024mobilellm,
    title={MobileLLM: Optimizing Sub-billion Parameter Language Models for On-Device Use Cases},
    author={Liu, Zechun and Zhao, Changsheng and Iandola, Forrest and Lai, Chen and Tian, Yuandong and Fedorov, Igor and Xiong, Yunyang and Chang, Ernie and Shi, Yangyang and Krishnamoorthi, Raghuraman and others},
    journal={arXiv preprint arXiv:2402.14905},
    year={2024}
}

@misc{acun2025composers,
      title={Composer: A Search Framework for Hybrid Neural Architecture Design}, 
      author={Bilge Acun and Prasoon Sinha and Newsha Ardalani and Sangmin Bae and Alicia Golden and Chien-Yu Lin and Meghana Madhyastha and Fei Sun and Neeraja J. Yadwadkar and Carole-Jean Wu},
      year={2025},
      eprint={2510.00379},
      archivePrefix={arXiv},
      primaryClass={cs.LG},
      url={https://arxiv.org/abs/2510.00379}, 
}

@inproceedings{yang2025gated,
title={Gated Delta Networks: Improving Mamba2 with Delta Rule},
author={Songlin Yang and Jan Kautz and Ali Hatamizadeh},
booktitle={The Thirteenth International Conference on Learning Representations},
year={2025},
url={https://openreview.net/forum?id=r8H7xhYPwz}
}

@inproceedings{mamba2,
author = {Dao, Tri and Gu, Albert},
title = {Transformers are SSMs: generalized models and efficient algorithms through structured state space duality},
year = {2024},
publisher = {JMLR.org},
booktitle = {Proceedings of the 41st International Conference on Machine Learning},
articleno = {399},
numpages = {31},
location = {Vienna, Austria},
series = {ICML'24}
}

@inproceedings{
lenz2025jamba,
title={Jamba: Hybrid Transformer-Mamba Language Models},
author={Barak Lenz and Opher Lieber and Alan Arazi and Amir Bergman and Avshalom Manevich and Barak Peleg and Ben Aviram and Chen Almagor and Clara Fridman and Dan Padnos and Daniel Gissin and Daniel Jannai and Dor Muhlgay and Dor Zimberg and Edden M. Gerber and Elad Dolev and Eran Krakovsky and Erez Safahi and Erez Schwartz and Gal Cohen and Gal Shachaf and Haim Rozenblum and Hofit Bata and Ido Blass and Inbal Magar and Itay Dalmedigos and Jhonathan Osin and Julie Fadlon and Maria Rozman and Matan Danos and Michael Gokhman and Mor Zusman and Naama Gidron and Nir Ratner and Noam Gat and Noam Rozen and Oded Fried and Ohad Leshno and Omer Antverg and Omri Abend and Or Dagan and Orit Cohavi and Raz Alon and Ro'i Belson and Roi Cohen and Rom Gilad and Roman Glozman and Shahar Lev and Shai Shalev-Shwartz and Shaked Haim Meirom and Tal Delbari and Tal Ness and Tomer Asida and Tom Ben Gal and Tom Braude and Uriya Pumerantz and Josh Cohen and Yonatan Belinkov and Yuval Globerson and Yuval Peleg Levy and Yoav Shoham},
booktitle={The Thirteenth International Conference on Learning Representations},
year={2025},
url={https://openreview.net/forum?id=JFPaD7lpBD}
}

@misc{glorioso2024zambacompact7bssm,
      title={Zamba: A Compact 7B SSM Hybrid Model}, 
      author={Paolo Glorioso and Quentin Anthony and Yury Tokpanov and James Whittington and Jonathan Pilault and Adam Ibrahim and Beren Millidge},
      year={2024},
      eprint={2405.16712},
      archivePrefix={arXiv},
      primaryClass={cs.LG},
      url={https://arxiv.org/abs/2405.16712}, 
}

@inproceedings{
ren2025samba,
title={Samba: Simple Hybrid State Space Models for Efficient Unlimited Context Language Modeling},
author={Liliang Ren and Yang Liu and Yadong Lu and yelong shen and Chen Liang and Weizhu Chen},
booktitle={The Thirteenth International Conference on Learning Representations},
year={2025},
url={https://openreview.net/forum?id=bIlnpVM4bc}
}

@inproceedings{
pilault2023blockstate,
title={Block-State Transformers},
author={Jonathan Pilault and Mahan Fathi and Orhan Firat and Christopher Pal and Pierre-Luc Bacon and Ross Goroshin},
booktitle={Thirty-seventh Conference on Neural Information Processing Systems},
year={2023},
url={https://openreview.net/forum?id=XRTxIBs2eu}
}

@misc{de2024griffinmixinggatedlinear,
      title={Griffin: Mixing Gated Linear Recurrences with Local Attention for Efficient Language Models}, 
      author={Soham De and Samuel L. Smith and Anushan Fernando and Aleksandar Botev and George Cristian-Muraru and Albert Gu and Ruba Haroun and Leonard Berrada and Yutian Chen and Srivatsan Srinivasan and Guillaume Desjardins and Arnaud Doucet and David Budden and Yee Whye Teh and Razvan Pascanu and Nando De Freitas and Caglar Gulcehre},
      year={2024},
      eprint={2402.19427},
      archivePrefix={arXiv},
      primaryClass={cs.LG},
      url={https://arxiv.org/abs/2402.19427}, 
}

@misc{child2019generatinglongsequencessparse,
      title={Generating Long Sequences with Sparse Transformers}, 
      author={Rewon Child and Scott Gray and Alec Radford and Ilya Sutskever},
      year={2019},
      eprint={1904.10509},
      archivePrefix={arXiv},
      primaryClass={cs.LG},
      url={https://arxiv.org/abs/1904.10509}, 
}

@article{kendall_tau,
 ISSN = {00063444},
 URL = {http://www.jstor.org/stable/2332226},
 author = {M. G. Kendall},
 journal = {Biometrika},
 number = {1/2},
 pages = {81--93},
 publisher = {[Oxford University Press, Biometrika Trust]},
 title = {A New Measure of Rank Correlation},
 urldate = {2026-02-13},
 volume = {30},
 year = {1938}
}

@misc{ba2016layernormalization,
      title={Layer Normalization}, 
      author={Jimmy Lei Ba and Jamie Ryan Kiros and Geoffrey E. Hinton},
      year={2016},
      eprint={1607.06450},
      archivePrefix={arXiv},
      primaryClass={stat.ML},
      url={https://arxiv.org/abs/1607.06450}, 
}

@inproceedings{
olson2025ax,
title={Ax: A Platform for Adaptive Experimentation},
author={Miles Olson and Elizabeth Santorella and Louis C. Tiao and Sait Cakmak and David Eriksson and Mia Garrard and Sam Daulton and Maximilian Balandat and Eytan Bakshy and Elena Kashtelyan and Zhiyuan Jerry Lin and Sebastian Ament and Bernard Beckerman and Eric Onofrey and Paschal Igusti and Cristian Lara and Benjamin Letham and Cesar Cardoso and Shiyun Sunny Shen and Andy Chenyuan Lin and Matthew Grange},
booktitle={AutoML 2025 ABCD Track},
year={2025},
url={https://openreview.net/forum?id=U1f6wHtG1g}
}

@inproceedings{
eriksson2021latencyaware,
title={Latency-Aware Neural Architecture Search with Multi-Objective Bayesian Optimization},
author={David Eriksson and Pierce I-Jen Chuang and Samuel Daulton and Peng Xia and Akshat Shrivastava and Arun Babu and Shicong Zhao and Ahmed A Aly and Ganesh Venkatesh and Maximilian Balandat},
booktitle={8th ICML Workshop on Automated Machine Learning (AutoML) },
year={2021},
url={https://openreview.net/forum?id=0ciyfd4SvbI}
}

@article{daulton2021qNEHVI,
  author       = {Samuel Daulton and
                  Maximilian Balandat and
                  Eytan Bakshy},
  title        = {Parallel Bayesian Optimization of Multiple Noisy Objectives with Expected
                  Hypervolume Improvement},
  journal      = {CoRR},
  volume       = {abs/2105.08195},
  year         = {2021},
  url          = {https://arxiv.org/abs/2105.08195},
  eprinttype    = {arXiv},
  eprint       = {2105.08195},
  bibsource    = {dblp computer science bibliography, https://dblp.org}
}

@misc{gemmateam2025gemma3technicalreport,
      title={Gemma 3 Technical Report}, 
      author={Gemma Team and Aishwarya Kamath and Johan Ferret and Shreya Pathak and Nino Vieillard and Ramona Merhej and Sarah Perrin and Tatiana Matejovicova and Alexandre Ramé and Morgane Rivière and Louis Rouillard and Thomas Mesnard and Geoffrey Cideron and Jean-bastien Grill and Sabela Ramos and Edouard Yvinec and Michelle Casbon and Etienne Pot and Ivo Penchev and Gaël Liu and Francesco Visin and Kathleen Kenealy and Lucas Beyer and Xiaohai Zhai and Anton Tsitsulin and Robert Busa-Fekete and Alex Feng and Noveen Sachdeva and Benjamin Coleman and Yi Gao and Basil Mustafa and Iain Barr and Emilio Parisotto and David Tian and Matan Eyal and Colin Cherry and Jan-Thorsten Peter and Danila Sinopalnikov and Surya Bhupatiraju and Rishabh Agarwal and Mehran Kazemi and Dan Malkin and Ravin Kumar and David Vilar and Idan Brusilovsky and Jiaming Luo and Andreas Steiner and Abe Friesen and Abhanshu Sharma and Abheesht Sharma and Adi Mayrav Gilady and Adrian Goedeckemeyer and Alaa Saade and Alex Feng and Alexander Kolesnikov and Alexei Bendebury and Alvin Abdagic and Amit Vadi and András György and André Susano Pinto and Anil Das and Ankur Bapna and Antoine Miech and Antoine Yang and Antonia Paterson and Ashish Shenoy and Ayan Chakrabarti and Bilal Piot and Bo Wu and Bobak Shahriari and Bryce Petrini and Charlie Chen and Charline Le Lan and Christopher A. Choquette-Choo and CJ Carey and Cormac Brick and Daniel Deutsch and Danielle Eisenbud and Dee Cattle and Derek Cheng and Dimitris Paparas and Divyashree Shivakumar Sreepathihalli and Doug Reid and Dustin Tran and Dustin Zelle and Eric Noland and Erwin Huizenga and Eugene Kharitonov and Frederick Liu and Gagik Amirkhanyan and Glenn Cameron and Hadi Hashemi and Hanna Klimczak-Plucińska and Harman Singh and Harsh Mehta and Harshal Tushar Lehri and Hussein Hazimeh and Ian Ballantyne and Idan Szpektor and Ivan Nardini and Jean Pouget-Abadie and Jetha Chan and Joe Stanton and John Wieting and Jonathan Lai and Jordi Orbay and Joseph Fernandez and Josh Newlan and Ju-yeong Ji and Jyotinder Singh and Kat Black and Kathy Yu and Kevin Hui and Kiran Vodrahalli and Klaus Greff and Linhai Qiu and Marcella Valentine and Marina Coelho and Marvin Ritter and Matt Hoffman and Matthew Watson and Mayank Chaturvedi and Michael Moynihan and Min Ma and Nabila Babar and Natasha Noy and Nathan Byrd and Nick Roy and Nikola Momchev and Nilay Chauhan and Noveen Sachdeva and Oskar Bunyan and Pankil Botarda and Paul Caron and Paul Kishan Rubenstein and Phil Culliton and Philipp Schmid and Pier Giuseppe Sessa and Pingmei Xu and Piotr Stanczyk and Pouya Tafti and Rakesh Shivanna and Renjie Wu and Renke Pan and Reza Rokni and Rob Willoughby and Rohith Vallu and Ryan Mullins and Sammy Jerome and Sara Smoot and Sertan Girgin and Shariq Iqbal and Shashir Reddy and Shruti Sheth and Siim Põder and Sijal Bhatnagar and Sindhu Raghuram Panyam and Sivan Eiger and Susan Zhang and Tianqi Liu and Trevor Yacovone and Tyler Liechty and Uday Kalra and Utku Evci and Vedant Misra and Vincent Roseberry and Vlad Feinberg and Vlad Kolesnikov and Woohyun Han and Woosuk Kwon and Xi Chen and Yinlam Chow and Yuvein Zhu and Zichuan Wei and Zoltan Egyed and Victor Cotruta and Minh Giang and Phoebe Kirk and Anand Rao and Kat Black and Nabila Babar and Jessica Lo and Erica Moreira and Luiz Gustavo Martins and Omar Sanseviero and Lucas Gonzalez and Zach Gleicher and Tris Warkentin and Vahab Mirrokni and Evan Senter and Eli Collins and Joelle Barral and Zoubin Ghahramani and Raia Hadsell and Yossi Matias and D. Sculley and Slav Petrov and Noah Fiedel and Noam Shazeer and Oriol Vinyals and Jeff Dean and Demis Hassabis and Koray Kavukcuoglu and Clement Farabet and Elena Buchatskaya and Jean-Baptiste Alayrac and Rohan Anil and Dmitry and Lepikhin and Sebastian Borgeaud and Olivier Bachem and Armand Joulin and Alek Andreev and Cassidy Hardin and Robert Dadashi and Léonard Hussenot},
      year={2025},
      eprint={2503.19786},
      archivePrefix={arXiv},
      primaryClass={cs.CL},
      url={https://arxiv.org/abs/2503.19786}, 
}

@misc{grattafiori2024llama3herdmodels,
      title={The Llama 3 Herd of Models}, 
      author={Aaron Grattafiori and Abhimanyu Dubey and Abhinav Jauhri and Abhinav Pandey and Abhishek Kadian and Ahmad Al-Dahle and Aiesha Letman and Akhil Mathur and Alan Schelten and Alex Vaughan and Amy Yang and Angela Fan and Anirudh Goyal and Anthony Hartshorn and Aobo Yang and Archi Mitra and Archie Sravankumar and Artem Korenev and Arthur Hinsvark and Arun Rao and Aston Zhang and Aurelien Rodriguez and Austen Gregerson and Ava Spataru and Baptiste Roziere and Bethany Biron and Binh Tang and Bobbie Chern and Charlotte Caucheteux and Chaya Nayak and Chloe Bi and Chris Marra and Chris McConnell and Christian Keller and Christophe Touret and Chunyang Wu and Corinne Wong and Cristian Canton Ferrer and Cyrus Nikolaidis and Damien Allonsius and Daniel Song and Danielle Pintz and Danny Livshits and Danny Wyatt and David Esiobu and Dhruv Choudhary and Dhruv Mahajan and Diego Garcia-Olano and Diego Perino and Dieuwke Hupkes and Egor Lakomkin and Ehab AlBadawy and Elina Lobanova and Emily Dinan and Eric Michael Smith and Filip Radenovic and Francisco Guzmán and Frank Zhang and Gabriel Synnaeve and Gabrielle Lee and Georgia Lewis Anderson and Govind Thattai and Graeme Nail and Gregoire Mialon and Guan Pang and Guillem Cucurell and Hailey Nguyen and Hannah Korevaar and Hu Xu and Hugo Touvron and Iliyan Zarov and Imanol Arrieta Ibarra and Isabel Kloumann and Ishan Misra and Ivan Evtimov and Jack Zhang and Jade Copet and Jaewon Lee and Jan Geffert and Jana Vranes and Jason Park and Jay Mahadeokar and Jeet Shah and Jelmer van der Linde and Jennifer Billock and Jenny Hong and Jenya Lee and Jeremy Fu and Jianfeng Chi and Jianyu Huang and Jiawen Liu and Jie Wang and Jiecao Yu and Joanna Bitton and Joe Spisak and Jongsoo Park and Joseph Rocca and Joshua Johnstun and Joshua Saxe and Junteng Jia and Kalyan Vasuden Alwala and Karthik Prasad and Kartikeya Upasani and Kate Plawiak and Ke Li and Kenneth Heafield and Kevin Stone and Khalid El-Arini and Krithika Iyer and Kshitiz Malik and Kuenley Chiu and Kunal Bhalla and Kushal Lakhotia and Lauren Rantala-Yeary and Laurens van der Maaten and Lawrence Chen and Liang Tan and Liz Jenkins and Louis Martin and Lovish Madaan and Lubo Malo and Lukas Blecher and Lukas Landzaat and Luke de Oliveira and Madeline Muzzi and Mahesh Pasupuleti and Mannat Singh and Manohar Paluri and Marcin Kardas and Maria Tsimpoukelli and Mathew Oldham and Mathieu Rita and Maya Pavlova and Melanie Kambadur and Mike Lewis and Min Si and Mitesh Kumar Singh and Mona Hassan and Naman Goyal and Narjes Torabi and Nikolay Bashlykov and Nikolay Bogoychev and Niladri Chatterji and Ning Zhang and Olivier Duchenne and Onur Çelebi and Patrick Alrassy and Pengchuan Zhang and Pengwei Li and Petar Vasic and Peter Weng and Prajjwal Bhargava and Pratik Dubal and Praveen Krishnan and Punit Singh Koura and Puxin Xu and Qing He and Qingxiao Dong and Ragavan Srinivasan and Raj Ganapathy and Ramon Calderer and Ricardo Silveira Cabral and Robert Stojnic and Roberta Raileanu and Rohan Maheswari and Rohit Girdhar and Rohit Patel and Romain Sauvestre and Ronnie Polidoro and Roshan Sumbaly and Ross Taylor and Ruan Silva and Rui Hou and Rui Wang and Saghar Hosseini and Sahana Chennabasappa and Sanjay Singh and Sean Bell and Seohyun Sonia Kim and Sergey Edunov and Shaoliang Nie and Sharan Narang and Sharath Raparthy and Sheng Shen and Shengye Wan and Shruti Bhosale and Shun Zhang and Simon Vandenhende and Soumya Batra and Spencer Whitman and Sten Sootla and Stephane Collot and Suchin Gururangan and Sydney Borodinsky and Tamar Herman and Tara Fowler and Tarek Sheasha and Thomas Georgiou and Thomas Scialom and Tobias Speckbacher and Todor Mihaylov and Tong Xiao and Ujjwal Karn and Vedanuj Goswami and Vibhor Gupta and Vignesh Ramanathan and Viktor Kerkez and Vincent Gonguet and Virginie Do and Vish Vogeti and Vítor Albiero and Vladan Petrovic and Weiwei Chu and Wenhan Xiong and Wenyin Fu and Whitney Meers and Xavier Martinet and Xiaodong Wang and Xiaofang Wang and Xiaoqing Ellen Tan and Xide Xia and Xinfeng Xie and Xuchao Jia and Xuewei Wang and Yaelle Goldschlag and Yashesh Gaur and Yasmine Babaei and Yi Wen and Yiwen Song and Yuchen Zhang and Yue Li and Yuning Mao and Zacharie Delpierre Coudert and Zheng Yan and Zhengxing Chen and Zoe Papakipos and Aaditya Singh and Aayushi Srivastava and Abha Jain and Adam Kelsey and Adam Shajnfeld and Adithya Gangidi and Adolfo Victoria and Ahuva Goldstand and Ajay Menon and Ajay Sharma and Alex Boesenberg and Alexei Baevski and Allie Feinstein and Amanda Kallet and Amit Sangani and Amos Teo and Anam Yunus and Andrei Lupu and Andres Alvarado and Andrew Caples and Andrew Gu and Andrew Ho and Andrew Poulton and Andrew Ryan and Ankit Ramchandani and Annie Dong and Annie Franco and Anuj Goyal and Aparajita Saraf and Arkabandhu Chowdhury and Ashley Gabriel and Ashwin Bharambe and Assaf Eisenman and Azadeh Yazdan and Beau James and Ben Maurer and Benjamin Leonhardi and Bernie Huang and Beth Loyd and Beto De Paola and Bhargavi Paranjape and Bing Liu and Bo Wu and Boyu Ni and Braden Hancock and Bram Wasti and Brandon Spence and Brani Stojkovic and Brian Gamido and Britt Montalvo and Carl Parker and Carly Burton and Catalina Mejia and Ce Liu and Changhan Wang and Changkyu Kim and Chao Zhou and Chester Hu and Ching-Hsiang Chu and Chris Cai and Chris Tindal and Christoph Feichtenhofer and Cynthia Gao and Damon Civin and Dana Beaty and Daniel Kreymer and Daniel Li and David Adkins and David Xu and Davide Testuggine and Delia David and Devi Parikh and Diana Liskovich and Didem Foss and Dingkang Wang and Duc Le and Dustin Holland and Edward Dowling and Eissa Jamil and Elaine Montgomery and Eleonora Presani and Emily Hahn and Emily Wood and Eric-Tuan Le and Erik Brinkman and Esteban Arcaute and Evan Dunbar and Evan Smothers and Fei Sun and Felix Kreuk and Feng Tian and Filippos Kokkinos and Firat Ozgenel and Francesco Caggioni and Frank Kanayet and Frank Seide and Gabriela Medina Florez and Gabriella Schwarz and Gada Badeer and Georgia Swee and Gil Halpern and Grant Herman and Grigory Sizov and Guangyi and Zhang and Guna Lakshminarayanan and Hakan Inan and Hamid Shojanazeri and Han Zou and Hannah Wang and Hanwen Zha and Haroun Habeeb and Harrison Rudolph and Helen Suk and Henry Aspegren and Hunter Goldman and Hongyuan Zhan and Ibrahim Damlaj and Igor Molybog and Igor Tufanov and Ilias Leontiadis and Irina-Elena Veliche and Itai Gat and Jake Weissman and James Geboski and James Kohli and Janice Lam and Japhet Asher and Jean-Baptiste Gaya and Jeff Marcus and Jeff Tang and Jennifer Chan and Jenny Zhen and Jeremy Reizenstein and Jeremy Teboul and Jessica Zhong and Jian Jin and Jingyi Yang and Joe Cummings and Jon Carvill and Jon Shepard and Jonathan McPhie and Jonathan Torres and Josh Ginsburg and Junjie Wang and Kai Wu and Kam Hou U and Karan Saxena and Kartikay Khandelwal and Katayoun Zand and Kathy Matosich and Kaushik Veeraraghavan and Kelly Michelena and Keqian Li and Kiran Jagadeesh and Kun Huang and Kunal Chawla and Kyle Huang and Lailin Chen and Lakshya Garg and Lavender A and Leandro Silva and Lee Bell and Lei Zhang and Liangpeng Guo and Licheng Yu and Liron Moshkovich and Luca Wehrstedt and Madian Khabsa and Manav Avalani and Manish Bhatt and Martynas Mankus and Matan Hasson and Matthew Lennie and Matthias Reso and Maxim Groshev and Maxim Naumov and Maya Lathi and Meghan Keneally and Miao Liu and Michael L. Seltzer and Michal Valko and Michelle Restrepo and Mihir Patel and Mik Vyatskov and Mikayel Samvelyan and Mike Clark and Mike Macey and Mike Wang and Miquel Jubert Hermoso and Mo Metanat and Mohammad Rastegari and Munish Bansal and Nandhini Santhanam and Natascha Parks and Natasha White and Navyata Bawa and Nayan Singhal and Nick Egebo and Nicolas Usunier and Nikhil Mehta and Nikolay Pavlovich Laptev and Ning Dong and Norman Cheng and Oleg Chernoguz and Olivia Hart and Omkar Salpekar and Ozlem Kalinli and Parkin Kent and Parth Parekh and Paul Saab and Pavan Balaji and Pedro Rittner and Philip Bontrager and Pierre Roux and Piotr Dollar and Polina Zvyagina and Prashant Ratanchandani and Pritish Yuvraj and Qian Liang and Rachad Alao and Rachel Rodriguez and Rafi Ayub and Raghotham Murthy and Raghu Nayani and Rahul Mitra and Rangaprabhu Parthasarathy and Raymond Li and Rebekkah Hogan and Robin Battey and Rocky Wang and Russ Howes and Ruty Rinott and Sachin Mehta and Sachin Siby and Sai Jayesh Bondu and Samyak Datta and Sara Chugh and Sara Hunt and Sargun Dhillon and Sasha Sidorov and Satadru Pan and Saurabh Mahajan and Saurabh Verma and Seiji Yamamoto and Sharadh Ramaswamy and Shaun Lindsay and Shaun Lindsay and Sheng Feng and Shenghao Lin and Shengxin Cindy Zha and Shishir Patil and Shiva Shankar and Shuqiang Zhang and Shuqiang Zhang and Sinong Wang and Sneha Agarwal and Soji Sajuyigbe and Soumith Chintala and Stephanie Max and Stephen Chen and Steve Kehoe and Steve Satterfield and Sudarshan Govindaprasad and Sumit Gupta and Summer Deng and Sungmin Cho and Sunny Virk and Suraj Subramanian and Sy Choudhury and Sydney Goldman and Tal Remez and Tamar Glaser and Tamara Best and Thilo Koehler and Thomas Robinson and Tianhe Li and Tianjun Zhang and Tim Matthews and Timothy Chou and Tzook Shaked and Varun Vontimitta and Victoria Ajayi and Victoria Montanez and Vijai Mohan and Vinay Satish Kumar and Vishal Mangla and Vlad Ionescu and Vlad Poenaru and Vlad Tiberiu Mihailescu and Vladimir Ivanov and Wei Li and Wenchen Wang and Wenwen Jiang and Wes Bouaziz and Will Constable and Xiaocheng Tang and Xiaojian Wu and Xiaolan Wang and Xilun Wu and Xinbo Gao and Yaniv Kleinman and Yanjun Chen and Ye Hu and Ye Jia and Ye Qi and Yenda Li and Yilin Zhang and Ying Zhang and Yossi Adi and Youngjin Nam and Yu and Wang and Yu Zhao and Yuchen Hao and Yundi Qian and Yunlu Li and Yuzi He and Zach Rait and Zachary DeVito and Zef Rosnbrick and Zhaoduo Wen and Zhenyu Yang and Zhiwei Zhao and Zhiyu Ma},
      year={2024},
      eprint={2407.21783},
      archivePrefix={arXiv},
      primaryClass={cs.AI},
      url={https://arxiv.org/abs/2407.21783}, 
}

@inproceedings{hellaswag,
    title = "{H}ella{S}wag: Can a Machine Really Finish Your Sentence?",
    author = "Zellers, Rowan  and
      Holtzman, Ari  and
      Bisk, Yonatan  and
      Farhadi, Ali  and
      Choi, Yejin",
    editor = "Korhonen, Anna  and
      Traum, David  and
      M{\`a}rquez, Llu{\'i}s",
    booktitle = "Proceedings of the 57th Annual Meeting of the Association for Computational Linguistics",
    month = jul,
    year = "2019",
    address = "Florence, Italy",
    publisher = "Association for Computational Linguistics",
    url = "https://aclanthology.org/P19-1472/",
    doi = "10.18653/v1/P19-1472",
    pages = "4791--4800"
}

@inproceedings{boolq,
    title = "{B}ool{Q}: Exploring the Surprising Difficulty of Natural Yes/No Questions",
    author = "Clark, Christopher  and
      Lee, Kenton  and
      Chang, Ming-Wei  and
      Kwiatkowski, Tom  and
      Collins, Michael  and
      Toutanova, Kristina",
    editor = "Burstein, Jill  and
      Doran, Christy  and
      Solorio, Thamar",
    booktitle = "Proceedings of the 2019 Conference of the North {A}merican Chapter of the Association for Computational Linguistics: Human Language Technologies, Volume 1 (Long and Short Papers)",
    month = jun,
    year = "2019",
    address = "Minneapolis, Minnesota",
    publisher = "Association for Computational Linguistics",
    url = "https://aclanthology.org/N19-1300/",
    doi = "10.18653/v1/N19-1300",
    pages = "2924--2936"
}

@misc{piqa,
      title={PIQA: Reasoning about Physical Commonsense in Natural Language}, 
      author={Yonatan Bisk and Rowan Zellers and Ronan Le Bras and Jianfeng Gao and Yejin Choi},
      year={2019},
      eprint={1911.11641},
      archivePrefix={arXiv},
      primaryClass={cs.CL},
      url={https://arxiv.org/abs/1911.11641}, 
}

@inproceedings{siqa,
    title = "Social {IQ}a: Commonsense Reasoning about Social Interactions",
    author = "Sap, Maarten  and
      Rashkin, Hannah  and
      Chen, Derek  and
      Le Bras, Ronan  and
      Choi, Yejin",
    editor = "Inui, Kentaro  and
      Jiang, Jing  and
      Ng, Vincent  and
      Wan, Xiaojun",
    booktitle = "Proceedings of the 2019 Conference on Empirical Methods in Natural Language Processing and the 9th International Joint Conference on Natural Language Processing (EMNLP-IJCNLP)",
    month = nov,
    year = "2019",
    address = "Hong Kong, China",
    publisher = "Association for Computational Linguistics",
    url = "https://aclanthology.org/D19-1454/",
    doi = "10.18653/v1/D19-1454",
    pages = "4463--4473"
}

@inproceedings{triviaqa,
    title = "{T}rivia{QA}: A Large Scale Distantly Supervised Challenge Dataset for Reading Comprehension",
    author = "Joshi, Mandar  and
      Choi, Eunsol  and
      Weld, Daniel  and
      Zettlemoyer, Luke",
    editor = "Barzilay, Regina  and
      Kan, Min-Yen",
    booktitle = "Proceedings of the 55th Annual Meeting of the Association for Computational Linguistics (Volume 1: Long Papers)",
    month = jul,
    year = "2017",
    address = "Vancouver, Canada",
    publisher = "Association for Computational Linguistics",
    url = "https://aclanthology.org/P17-1147/",
    doi = "10.18653/v1/P17-1147",
    pages = "1601--1611"
}

@article{nq,
    title = "Natural Questions: A Benchmark for Question Answering Research",
    author = "Kwiatkowski, Tom  and
      Palomaki, Jennimaria  and
      Redfield, Olivia  and
      Collins, Michael  and
      Parikh, Ankur  and
      Alberti, Chris  and
      Epstein, Danielle  and
      Polosukhin, Illia  and
      Devlin, Jacob  and
      Lee, Kenton  and
      Toutanova, Kristina  and
      Jones, Llion  and
      Kelcey, Matthew  and
      Chang, Ming-Wei  and
      Dai, Andrew M.  and
      Uszkoreit, Jakob  and
      Le, Quoc  and
      Petrov, Slav",
    editor = "Lee, Lillian  and
      Johnson, Mark  and
      Roark, Brian  and
      Nenkova, Ani",
    journal = "Transactions of the Association for Computational Linguistics",
    volume = "7",
    year = "2019",
    address = "Cambridge, MA",
    publisher = "MIT Press",
    url = "https://aclanthology.org/Q19-1026/",
    doi = "10.1162/tacl_a_00276",
    pages = "452--466"
}

@misc{arcc,
      title={Think you have Solved Question Answering? Try ARC, the AI2 Reasoning Challenge}, 
      author={Peter Clark and Isaac Cowhey and Oren Etzioni and Tushar Khot and Ashish Sabharwal and Carissa Schoenick and Oyvind Tafjord},
      year={2018},
      eprint={1803.05457},
      archivePrefix={arXiv},
      primaryClass={cs.AI},
      url={https://arxiv.org/abs/1803.05457}, 
}

@article{winogrande,
author = {Sakaguchi, Keisuke and Bras, Ronan Le and Bhagavatula, Chandra and Choi, Yejin},
title = {WinoGrande: an adversarial winograd schema challenge at scale},
year = {2021},
issue_date = {September 2021},
publisher = {Association for Computing Machinery},
address = {New York, NY, USA},
volume = {64},
number = {9},
issn = {0001-0782},
url = {https://doi.org/10.1145/3474381},
doi = {10.1145/3474381},
journal = {Commun. ACM},
month = aug,
pages = {99–106},
numpages = {8}
}

@misc{lm-eval-harness,
  author       = {Gao, Leo and Tow, Jonathan and Abbasi, Baber and Biderman, Stella and Black, Sid and DiPofi, Anthony and Foster, Charles and Golding, Laurence and Hsu, Jeffrey and Le Noac'h, Alain and Li, Haonan and McDonell, Kyle and Muennighoff, Niklas and Ociepa, Chris and Phang, Jason and Reynolds, Laria and Schoelkopf, Hailey and Skowron, Aviya and Sutawika, Lintang and Tang, Eric and Thite, Anish and Wang, Ben and Wang, Kevin and Zou, Andy},
  title        = {A framework for few-shot language model evaluation},
  month        = 12,
  year         = 2023,
  publisher    = {Zenodo},
  version      = {v0.4.0},
  doi          = {10.5281/zenodo.10256836},
  url          = {https://zenodo.org/records/10256836}
}

@InProceedings{pmlr-v202-daulton23a,
  title = 	 {Hypervolume Knowledge Gradient: A Lookahead Approach for Multi-Objective {B}ayesian Optimization with Partial Information},
  author =       {Daulton, Sam and Balandat, Maximilian and Bakshy, Eytan},
  booktitle = 	 {Proceedings of the 40th International Conference on Machine Learning},
  pages = 	 {7167--7204},
  year = 	 {2023},
  editor = 	 {Krause, Andreas and Brunskill, Emma and Cho, Kyunghyun and Engelhardt, Barbara and Sabato, Sivan and Scarlett, Jonathan},
  volume = 	 {202},
  series = 	 {Proceedings of Machine Learning Research},
  month = 	 {23--29 Jul},
  publisher =    {PMLR},
  url = 	 {https://proceedings.mlr.press/v202/daulton23a.html}
}

@misc{yang2025qwen3technicalreport,
      title={Qwen3 Technical Report}, 
      author={An Yang and Anfeng Li and Baosong Yang and Beichen Zhang and Binyuan Hui and Bo Zheng and Bowen Yu and Chang Gao and Chengen Huang and Chenxu Lv and Chujie Zheng and Dayiheng Liu and Fan Zhou and Fei Huang and Feng Hu and Hao Ge and Haoran Wei and Huan Lin and Jialong Tang and Jian Yang and Jianhong Tu and Jianwei Zhang and Jianxin Yang and Jiaxi Yang and Jing Zhou and Jingren Zhou and Junyang Lin and Kai Dang and Keqin Bao and Kexin Yang and Le Yu and Lianghao Deng and Mei Li and Mingfeng Xue and Mingze Li and Pei Zhang and Peng Wang and Qin Zhu and Rui Men and Ruize Gao and Shixuan Liu and Shuang Luo and Tianhao Li and Tianyi Tang and Wenbiao Yin and Xingzhang Ren and Xinyu Wang and Xinyu Zhang and Xuancheng Ren and Yang Fan and Yang Su and Yichang Zhang and Yinger Zhang and Yu Wan and Yuqiong Liu and Zekun Wang and Zeyu Cui and Zhenru Zhang and Zhipeng Zhou and Zihan Qiu},
      year={2025},
      eprint={2505.09388},
      archivePrefix={arXiv},
      primaryClass={cs.CL},
      url={https://arxiv.org/abs/2505.09388}, 
}

@inproceedings{
bercovich2025puzzle,
title={Puzzle: Distillation-Based {NAS} for Inference-Optimized {LLM}s},
author={Akhiad Bercovich and Tomer Ronen and Talor Abramovich and Nir Ailon and Nave Assaf and Mohammed Dabbah and Ido Galil and Amnon Geifman and Yonatan Geifman and Izhak Golan and Netanel Haber and Ehud Dov Karpas and Roi Koren and Itay Levy and Pavlo Molchanov and Shahar Mor and Zach Moshe and Najeeb Nabwani and Omri Puny and Ran Rubin and Itamar Schen and Ido Shahaf and Oren Tropp and Omer Ullman Argov and Ran Zilberstein and Ran El-Yaniv},
booktitle={Forty-second International Conference on Machine Learning},
year={2025},
url={https://openreview.net/forum?id=RY5MMBHRqo}
}

@misc{sanyal2026nanosdedgeefficientfoundation,
      title={NanoSD: Edge Efficient Foundation Model for Real Time Image Restoration}, 
      author={Subhajit Sanyal and Srinivas Soumitri Miriyala and Akshay Janardan Bankar and Manjunath Arveti and Sowmya Vajrala and Shreyas Pandith and Sravanth Kodavanti and Abhishek Ameta and Harshit and Amit Satish Unde},
      year={2026},
      eprint={2601.09823},
      archivePrefix={arXiv},
      primaryClass={cs.CV},
      url={https://arxiv.org/abs/2601.09823}, 
}

@article{SOBOL196786,
title = {On the distribution of points in a cube and the approximate evaluation of integrals},
journal = {USSR Computational Mathematics and Mathematical Physics},
volume = {7},
number = {4},
pages = {86-112},
year = {1967},
issn = {0041-5553},
doi = {https://doi.org/10.1016/0041-5553(67)90144-9},
url = {https://www.sciencedirect.com/science/article/pii/0041555367901449},
author = {I.M Sobol'}
}

@article{scaling-law,
  author       = {Jared Kaplan and
                  Sam McCandlish and
                  Tom Henighan and
                  Tom B. Brown and
                  Benjamin Chess and
                  Rewon Child and
                  Scott Gray and
                  Alec Radford and
                  Jeffrey Wu and
                  Dario Amodei},
  title        = {Scaling Laws for Neural Language Models},
  journal      = {CoRR},
  volume       = {abs/2001.08361},
  year         = {2020},
  url          = {https://arxiv.org/abs/2001.08361},
  eprinttype   = {arXiv},
  eprint       = {2001.08361},
  bibsource    = {dblp computer science bibliography, https://dblp.org}
}

@inproceedings{training-compute-optimal,
author = {Hoffmann, Jordan and Borgeaud, Sebastian and Mensch, Arthur and Buchatskaya, Elena and Cai, Trevor and Rutherford, Eliza and de Las Casas, Diego and Hendricks, Lisa Anne and Welbl, Johannes and Clark, Aidan and Hennigan, Tom and Noland, Eric and Millican, Katie and van den Driessche, George and Damoc, Bogdan and Guy, Aurelia and Osindero, Simon and Simonyan, Karen and Elsen, Erich and Vinyals, Oriol and Rae, Jack W. and Sifre, Laurent},
title = {Training compute-optimal large language models},
year = {2022},
isbn = {9781713871088},
publisher = {Curran Associates Inc.},
address = {Red Hook, NY, USA},
booktitle = {Proceedings of the 36th International Conference on Neural Information Processing Systems},
articleno = {2176},
numpages = {15},
location = {New Orleans, LA, USA},
series = {NIPS '22}
}

@misc{meta2025llama4,
  title={Llama 4 Maverick and Scout},
  author={{Meta AI}},
  year={2025},
  howpublished={\url{https://huggingface.co/meta-llama/Llama-4-Scout-17B-16E}}
}
